\documentclass[]{article}
\usepackage{proceed2e}

\usepackage{epsfig,times,amsmath,amsfonts}

\RequirePackage{natbib}
\bibpunct{(}{)}{;}{a}{,}{,}

\newenvironment{enumerate*}%
  {\begin{enumerate}%
    \setlength{\itemsep}{0pt}%
    \setlength{\parskip}{0pt}}%
  {\end{enumerate}}

\newenvironment{itemize*}%
  {\begin{itemize}%
    \setlength{\itemsep}{0pt}%
    \setlength{\parskip}{0pt}}%
  {\end{itemize}}

\title{Gaussian Process Structural Equation Models with Latent Variables}

\author{ 
{\bf Ricardo Silva} \\  
Department of Statistical Science\\
University College London\\
{\tt ricardo@stats.ucl.ac.uk}\\
\And 
{\bf Robert B. Gramacy}  \\ 
Statistical Laboratory\\
University of Cambridge\\
{\tt bobby@statslab.cam.ac.uk}\\
} 

\begin{document}

\maketitle

\begin{abstract}
  In a variety of disciplines such as social sciences, psychology,
  medicine and economics, the recorded data are considered to be noisy
  measurements of latent variables connected by some causal structure.
  This corresponds to a family of graphical models known as the
  structural equation model with latent variables. While linear
  non-Gaussian variants have been well-studied, inference in
  nonparametric structural equation models is still underdeveloped. We
  introduce a sparse Gaussian process parameterization that defines a
  non-linear structure connecting latent variables, unlike common
  formulations of Gaussian process latent variable models. The sparse
  parameterization is given a full Bayesian treatment without
  compromising Markov chain Monte Carlo efficiency. We compare the
  stability of the sampling procedure and the predictive ability of
  the model against the current practice.
\end{abstract}

\section{CONTRIBUTION}

A cornerstone principle of many disciplines is that observations are
noisy measurements of hidden variables of interest. This is particularly
prominent in fields such as social sciences, psychology, marketing and
medicine. For instance, data can come in the form of social and
economical indicators, answers to questionnaires in a medical exam or
marketing survey, and instrument readings such as fMRI scans.  Such
indicators are treated as measures of latent factors such as the
latent ability levels of a subject in a psychological study, or the
abstract level of democratization of a country. The literature on
structural equation models (SEMs) \citep{bart:08,bol:89} approaches such
problems with directed graphical models, where each node in the graph
is a noisy function of its parents. The goals of the analysis include
typical applications of latent variable models, such as projecting
points in a latent space (with confidence regions) for ranking,
clustering and visualization; density estimation; missing data
imputation; and causal inference
\citep{pearl:00,sgs:00}.

This paper introduces a nonparametric formulation of SEMs with hidden
nodes, where functions connecting latent variables are given a
Gaussian process prior. An efficient but flexible sparse formulation
is adopted. To the best of our knowledge, our contribution is the
first full Gaussian process treatment of SEMs with latent variables.

We assume that the model graphical structure is given.  Structural
model selection with latent variables is a complex topic which we will
not pursue here: a detailed discussion of model selection is left as
future work. \cite{aspa:09} and \cite{sil:06} discuss relevant issues.
Our goal is to be able to generate posterior distributions over
parameters and latent variables with scalable sampling procedures with
good mixing properties, while being competitive against non-sparse
Gaussian process models.

In Section \ref{sec:model}, we specify the likelihood function for our
structural equation models and its implications. In Section
\ref{sec:sparse}, we elaborate on priors, Bayesian learning, and a
sparse variation of the basic model which is able to handle larger
datasets. Section \ref{sec:mcmc} describes a Markov chain Monte Carlo
(MCMC) procedure. Section \ref{sec:experiments} evaluates the
usefulness of the model and the stability of the sampler in a set of
real-world SEM applications with comparisons to modern alternatives. 
Finally, in Section \ref{sec:related} we discuss related work.

\begin{figure*}[ht]
\begin{center}
\begin{tabular}{cc}
\epsfig{file=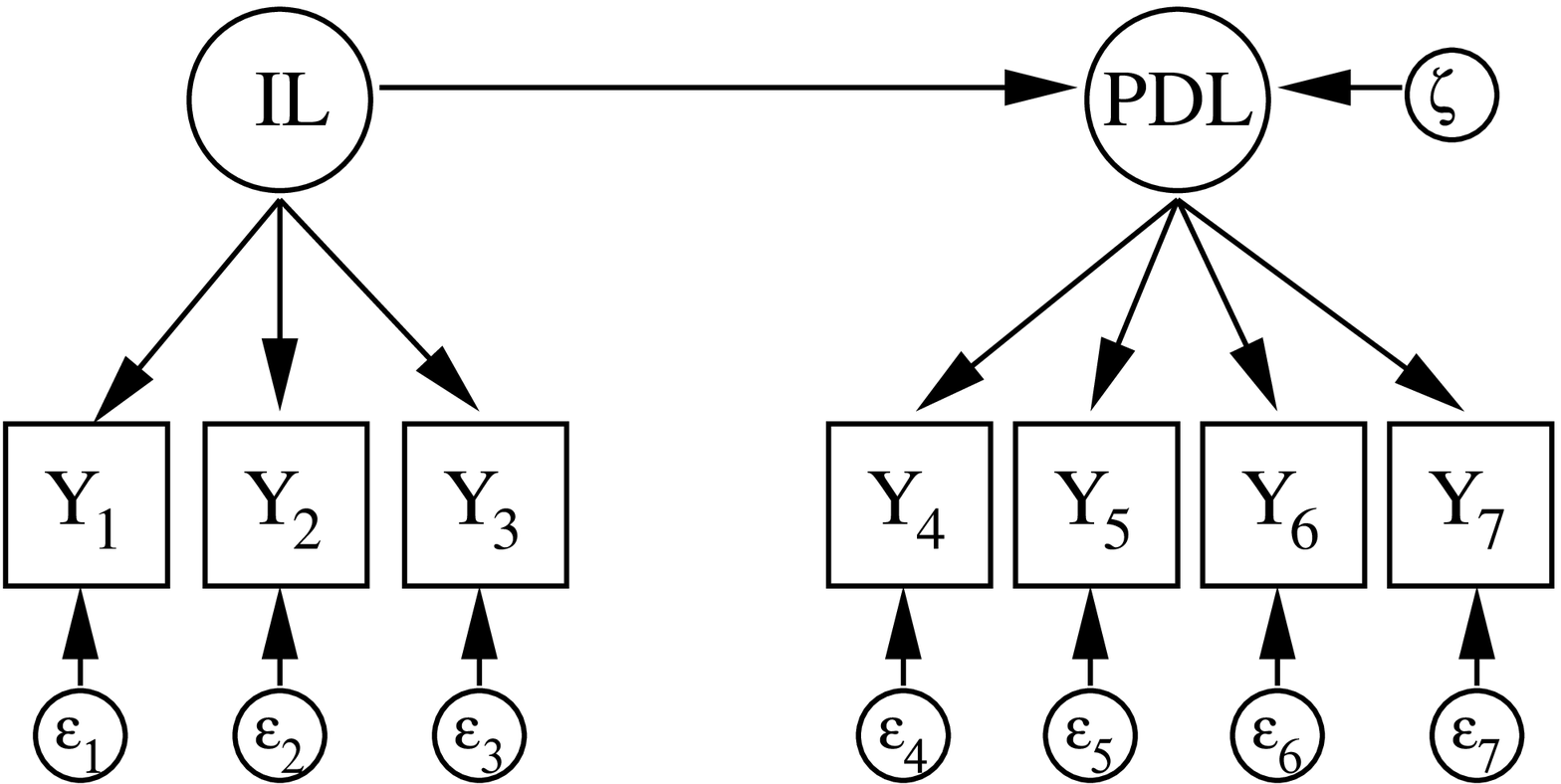,width=5.9cm}\hspace{0.3in} &
\epsfig{file=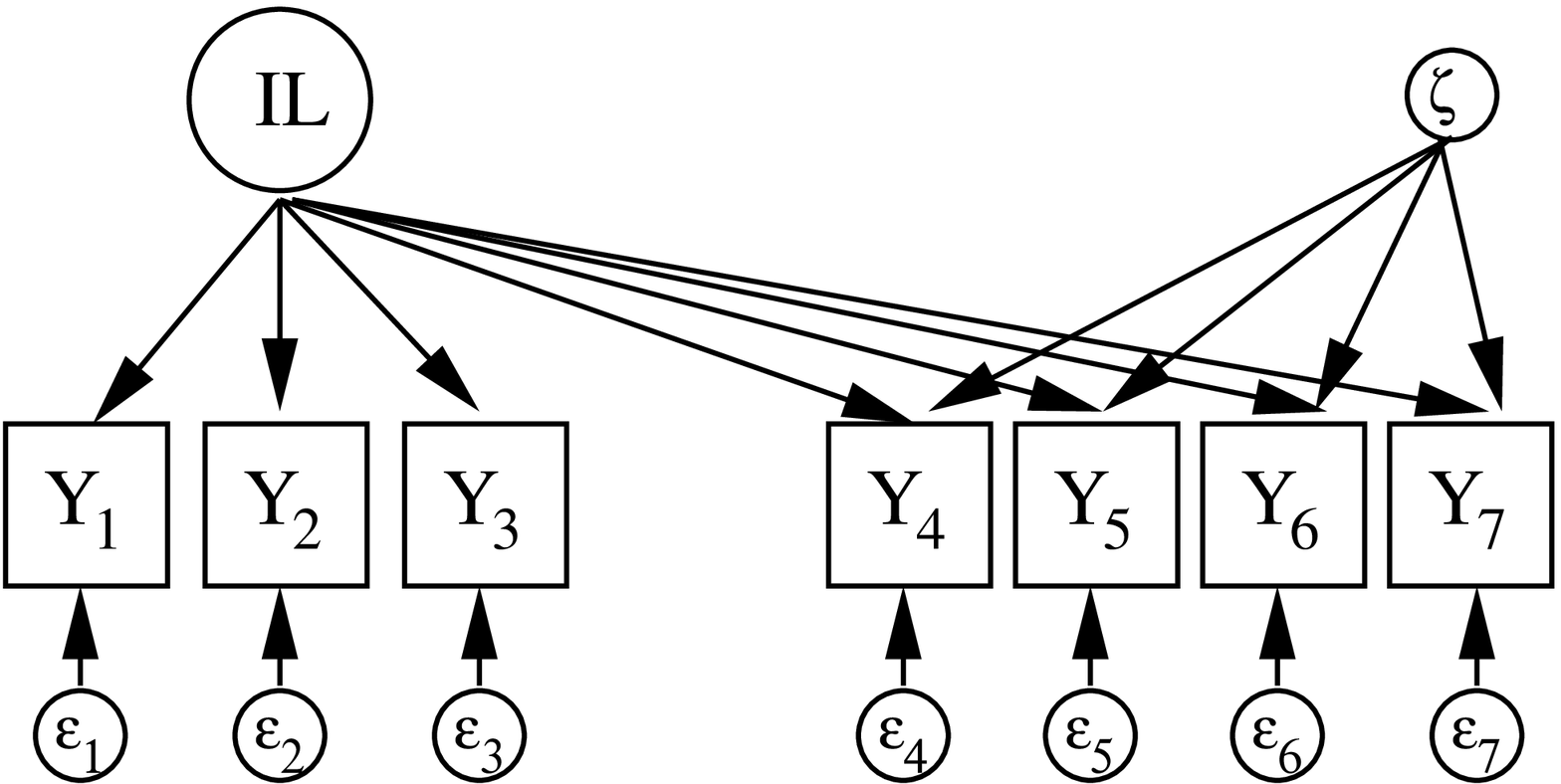,width=5.9cm} \\
(a) & (b)\\
\end{tabular}
\end{center}
\caption{(a) An example adapted from \cite{palomo:07}: latent variable $IL$ corresponds
to a scalar labeled as the industrialization level of a country.
$PDL$ is the corresponding political democratization level. Variables
$Y_1, Y_2, Y_3$ are indicators of industrialization (e.g., gross
national product) while $Y_4, \dots, Y_7$ are indicators of
democratization (e.g., expert assessements of freedom of
press). Each variable is a function of its parents with a
corresponding additive error term: $\epsilon_i$ for each $Y_i$, and
$\zeta$ for democratization levels. For instance, $PDL = f(IL) +
\zeta$ for some function $f(\cdot)$. (b) Dependence among latent
variables is essential to obtain sparsity in the measurement
structure. Here we depict how the graphical dependence structure would
look like if we regressed the observed variables on the independent
latent variables of (a).}
\label{fig:simple-bollen}
\end{figure*}

\section{THE MODEL: LIKELIHOOD}
\label{sec:model}

Let $\mathcal G$ be a given directed acyclic graph (DAG). For
simplicity, in this paper we assume that no observed variable is a
parent in $\mathcal G$ of any latent variable. Many SEM applications
are of this type \citep{bol:89,sil:06}, and this will simplify our
presentation. Likewise, we will treat models for continuous variables
only.  Although cyclic SEMs are also well-defined for the linear
case \citep{bol:89}, non-linear cyclic models are not trivial to define
and as such we will exclude them from this paper.

Let $\mathcal X$ be our set of latent variables and $X_i \in \mathcal
X$ be a particular latent variable. Let $\mathbf X_{P_i}$ be the set
of parents of $X_i$ in $\mathcal G$. The latent structure in our SEM is given by
the following generative model: if the parent set of $X_i$ is not empty,
\begin{equation}
\label{eq:m1}
X_i = f_i(\mathbf X_{P_i}) + \zeta_i, \text{ where }
\zeta_i \sim \mathcal N(0, v_{\zeta_i})
\end{equation}

\noindent $\mathcal N(m, v)$ is the Gaussian distribution with mean $m$ and variance
$v$. If $X_i$ has no parents (i.e., it is an {\bf exogenous} latent
variable, in SEM terminology), it is given a mixture of Gaussians
marginal\footnote{For simplicity of presentation, in this paper we
adopt a finite mixture of Gaussians marginal for the exogenous
variables. However, introducing a Dirichlet process mixture of
Gaussians marginal is conceptually straightforward.}.

The {\bf measurement model}, i.e., the model that describes the
distribution of observations $\mathcal Y$ given latent variables
$\mathcal X$, is as follows. For each $Y_j \in \mathcal Y$ with parent
set $\mathbf X_{P_j}$, we have
\begin{equation}
\label{eq:m2}
         Y_j  =    \lambda_{j0} + \mathbf X_{P_j}^{\mathsf T}\Lambda_j + \epsilon_j, 
\mbox{where }
  \epsilon_j  \sim  \mathcal{N}(0, v_{\epsilon_j})
\end{equation}

\noindent Error terms $\{\epsilon_j\}$ are assumed to be mutually independent and
independent of all latent variables in $\mathcal X$. Moreover, $\Lambda_j$ is a
vector of linear coefficients $\Lambda_j = [\lambda_{j1}\ \dots\
\lambda_{j|\mathbf X_{P_j}|}]^\mathsf{T}$.  Following SEM terminology,
we say that $Y_j$ is an {\bf indicator} of the latent variables in
$\mathbf X_{P_j}$.

An example is shown in Figure \ref{fig:simple-bollen}(a). Following
the notation of \cite{bol:89}, squares represent observed variables
and circles, latent variables. SEMs are graphical models with an
emphasis on sparse models where: 1. latent variables are dependent
according to a directed graph model; 2. observed variables measure
(i.e., are children of) very few latent variables. Although
sparse latent variable models have been the object of study in machine
learning and statistics (e.g., \cite{wood:06,zou:06}), not much has
been done on exploring nonparametric models with dependent latent
structure (a loosely related exception being dynamic systems, where
filtering is the typical application). Figure
\ref{fig:simple-bollen}(b) illustrates how modeling can be affected by
discarding the structure among latents\footnote{Another consequence
  of modeling latent dependencies is reducing the number of parameters
  of the model: a SEM with a linear measurement model can be seen as a
  type of module network \citep{segal:05} where the observed children
  of a particular latent $X_i$ share the same nonlinearities
  propagated from $\mathbf X_{P_i}$: in the context of Figure
  \ref{fig:simple-bollen}, each indicator $Y_i \in \{Y_4, \dots,
  Y_7\}$ has a conditional expected value of $\lambda_{i0} +
  \lambda_{i1}f_2(X_1)$ for a given $X_1$: function $f_2(\cdot)$ is
  shared among the indicators of $X_2$.}.

\subsection{Identifiability Conditions}
\label{sec:identifiability}

Latent variable models might be unidentifiable. In the context of
Bayesian inference, this is less of a theoretical issue than a computational
one: unidentifiable models might lead to poor mixing in MCMC,
as discussed in Section \ref{sec:experiments}. Moreover,
in many applications, the latent embedding of the data points is of
interest itself, or the latent regression functions are relevant for
causal inference purposes. In such applications, an unidentifiable
model is of limited interest. In this Section, we show how to 
derive sufficient conditions for identifiability.

Consider the case where a latent variable $X_i$ has at least three
unique indicators $\mathcal Y_i \equiv \{Y_{i\alpha}, Y_{i\beta},
Y_{i\gamma}\}$, in the sense that no element in $\mathcal Y_i$ has any
other parent in $\mathcal G$ but $X_i$.  It is known that in this case
\citep{bol:89} the parameters of the structural equations for
each element of $\mathcal Y_i$ are identifiable (i.e., the linear
coefficients and the error term variance) up to a scale and
sign of the latent variable. This can be resolved by setting the
linear structural equation of (say) $Y_{i\alpha}$ to $Y_{i\alpha} =
X_i + \epsilon_{i\alpha}$. The distribution of the error terms is then
identifiable. The distribution of $X_i$ follows from a
deconvolution between the observed distribution of an element of
$\mathcal Y_i$ and the identified distribution of the error
term.

Identifiability of the joint of $\mathcal X$ can be resolved by
multivariate deconvolution under extra assumptions. For instance,
\cite{masry:03} describes the problem in the context of kernel
density estimation (with known joint distribution of error terms, but
unknown joint of $\mathcal Y$).

Assumptions for the identifiability of functions $f_i(\cdot)$, given
the identifiability of the joint of $\mathcal X$, have been discussed
in the literature of error-in-variables regression
\citep{fan:93,carroll:04}. Error-in-variables regression is a special
case of our problem, where $X_i$ is observed but $\mathbf X_{P_i}$ is
not. However, since we have $Y_{i\alpha} = X_i + \epsilon_i$, this is
equivalent to a error-in-variables regression $Y_{i\alpha} =
f_i(\mathbf X_{P_i}) + \epsilon_{i\alpha} + \zeta_i$, where the
compound error term $\epsilon_{i\alpha} + \zeta_i$ is still
independent of $\mathbf X_{P_i}$.

It can be shown that such identifiability conditions can be exploited
in order to identify causal directionality among latent variables
under additional assumptions, as discussed by \cite{hoyer:08} for the
fully observed case\footnote{Notice that if the distribution of the
  error terms is non-Gaussian, identification is easier: we only need
  two unique indicators $Y_{i\alpha}$ and $Y_{i\beta}$: since
  $\epsilon_{i\alpha}, \epsilon_{i\beta}$ and $X_i$ are mutually
  independent, identification follows from known results derived in
  the literature of overcomplete independent component analysis
  \citep{hoyer:08b}.}. A brief discussion is presented in the Appendix.
In our context, we focus on the implications
of identifiabilty on MCMC (Section \ref{sec:experiments}).

\section{THE MODEL: PRIORS} 
\label{sec:sparse}

Each $f_i(\cdot)$ can be given a Gaussian process prior
\citep{rassmilliams:06}. In this case, we call this class of models the 
GPSEM-LV family, standing for Gaussian Process Structural Equation
Model with Latent Variables. Models without latent variables and
measurement models have been discussed by \cite{friedman:00b}\footnote{To see how
the Gaussian process networks of \cite{friedman:00b} are a special case of
GPSEM-LV, imagine a model where each latent variable is measured without error.
That is, each $X_i$ has at least one observed child $Y_i$ such that $Y_i = X_i$.
The measurement model is still linear, but each structural equation among
latent variables can be equivalently written in terms of the observed variables:
i.e., $X_i = f_i(\mathbf X_{P_i}) + \zeta_i$ is equivalent to 
$Y_i = f_i(\mathbf Y_{P_i}) + \zeta_i$, as in Friedman and Nachman.}.

\begin{figure*}[ht]
\begin{center}
\begin{tabular}{cc}
\epsfig{file=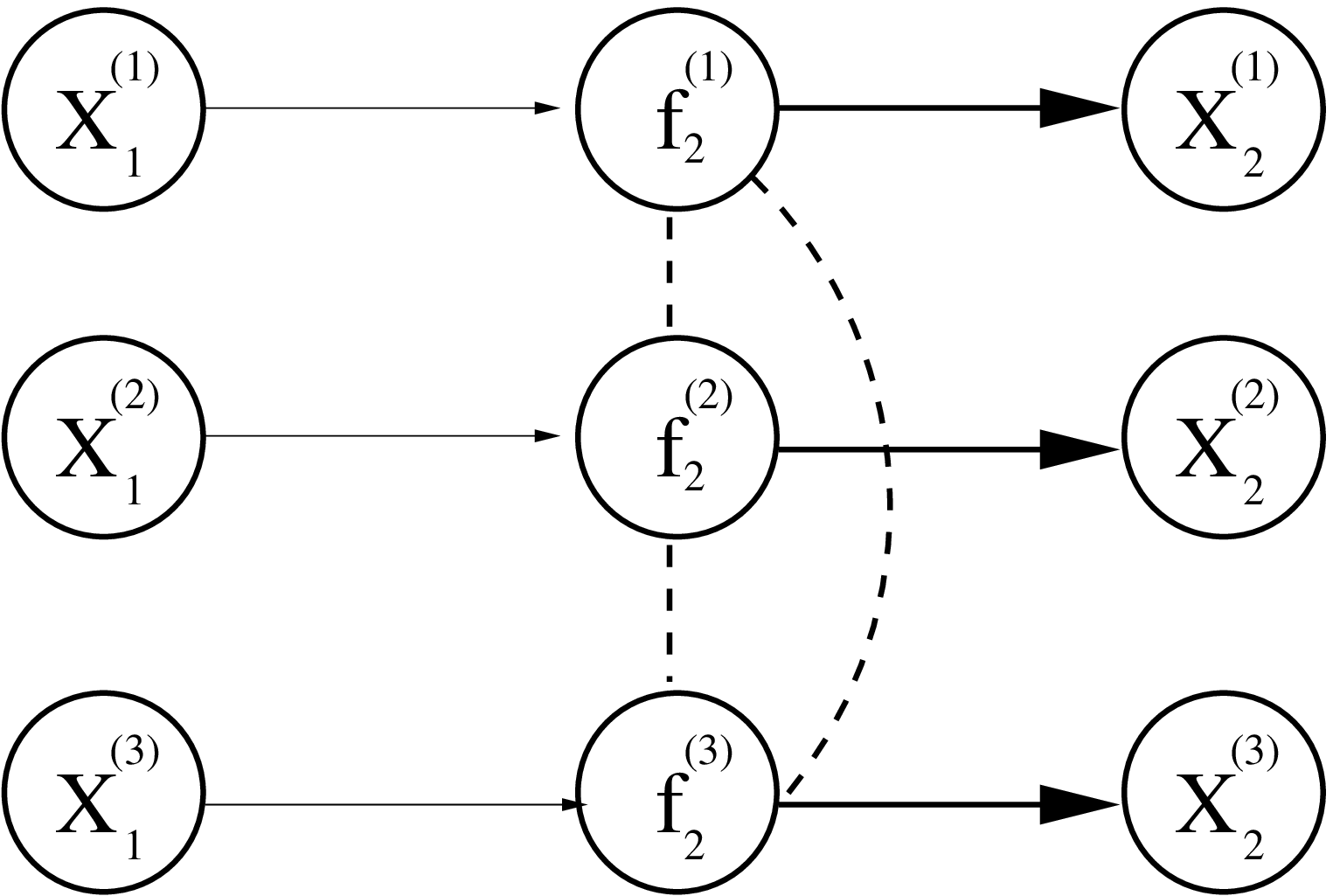,height=3.0cm}\hspace{0.3in} &
\epsfig{file=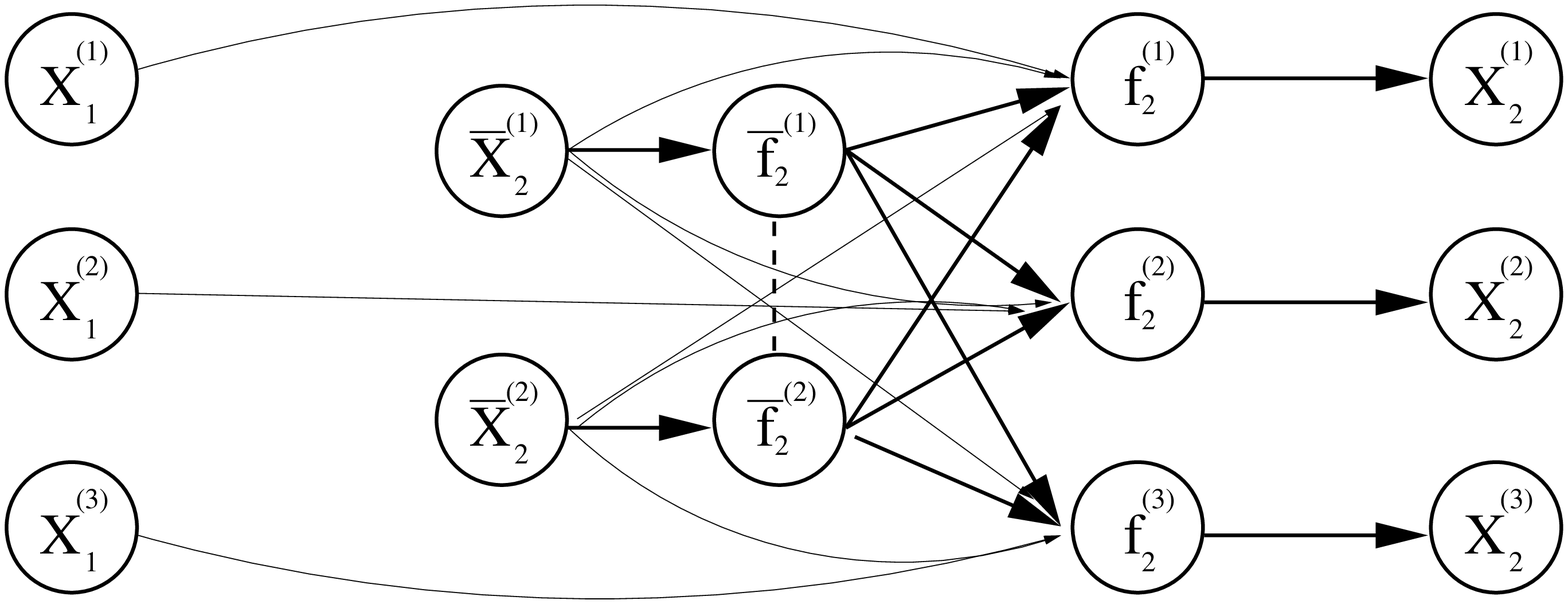,height=3.0cm} \\
(a) & (b)\\
\end{tabular}
\end{center}
\caption{(a) This figure depicts a model for the latent structure
$X_1 \rightarrow X_2$ with $N = 3$ (edges into latent functions
are lighter for visualization purposes only) using a standard Gaussian process prior. Dashed arrows
represent that function values $\{f_2^{(1)}, f_2^{(2)}, f_2^{(3)}\}$ are
mutually dependent even after conditioning on $\{X_1^{(1)}, X_1^{(2)},
X_1^{(3)}\}$. In (b), we have the graphical depiction of the respective
Bayesian pseudo-inputs model with $M = 2$. Althought the model is seemingly
more complex, it scales much better: mutual dependencies are confined to
the clique of pseudo-functions, which scales by $M$ instead.}
\label{fig:bspgp}
\end{figure*}

\subsection{Gaussian Process Prior and Notation}
\label{sec:pseudo-review}

Let $X_i$ be an arbitrary latent variable in the graph, with latent
parents $\mathbf X_{P_i}$. We will use $\mathbf X^{(d)}$ to represent
the $d^{th}$ $\mathbf X$ sampled from the distribution of random
vector $\mathbf X$, and $X_i^{(d)}$ indexes its $i^{th}$ component.
For instance, $\mathbf X_{P_i}^{(d)}$ is the $d^{th}$ sample of the parents
of $X_i$. A training set of size $N$ is represented as
$\{\mathbf Z^{(1)}, \dots, \mathbf Z^{(N)}\}$, where $\mathbf Z$ is
the set of all variables. Lower case $\mathbf x$ represents fixed
values of latent variables, and $\mathbf x^{1:N}$ represents a whole
set $\{\mathbf x^{(1)}, \dots, \mathbf x^{(N)}\}$.

For each $\mathbf x_{P_i}$, the corresponding Gaussian process prior
for function values $\mathbf f_i^{1:N} \equiv \{f_i^{(1)}, \dots, f_i^{(N)}\}$
is
\begin{equation*}
\mathbf f_i^{1:N}\ |\ \mathbf x_{P_i}^{1:N} \sim \mathcal N(0, \mathbf K_i)
\end{equation*}
\noindent where $\mathbf K_i$ is a $N \times N$ kernel matrix
\citep{rassmilliams:06}, as determined by $\mathbf x_{P_i}^{1:N}$.
Each corresponding $x_i^{(d)}$ is given by $f_i^{(d)} + \zeta_i^{(d)}$, as in
Equation (\ref{eq:m1}).

MCMC can be used to sample from the posterior distribution over latent
variables and functions. However, each sampling step in this model
costs $\mathcal O(N^3)$, making sampling very slow when $N$ is at the
order of hundreds, and essentially undoable when $N$ is in the
thousands. As an alternative, we introduce a multilayered
representation adapted from the pseudo-inputs model of
\cite{snelson:06}. The goal is to reduce the sampling cost down to
$\mathcal O(M^2N)$, $M < N$. $M$ can be chosen according to the available
computational resources.

\subsection{Pseudo-inputs Review}
\label{sec:pseudo-review}

We briefly review the pseudo-inputs model \citep{snelson:06} in our notation.
As before, let $\mathbf X^{(d)}$ represent the $d^{th}$ data point for some $\mathbf X$.
For a set $\mathbf X_i^{1:N} \equiv \{X_i^{(1)}, \dots,
X_i^{(N)}\}$ with corresponding parent set $\mathbf X_{P_i}^{1:N} \equiv \{\mathbf
X_{P_i}^{(1)}, \dots, \mathbf X_{P_i}^{(N)}\}$ and corresponding latent
function values $\mathbf f_i^{1:N}$, we define a {\it pseudo-input} set 
${\bar{\mathbf X}_i}^{1:M} \equiv \{{\bar{\mathbf X}_i}^{(1)}, \dots, {\bar{\mathbf X}_i}^{(M)}\}$ such that
\begin{align}
\mathbf f_i^{1:N}\ |\ \mathbf x_{P_i}^{1:N}, \bar{\mathbf f}_i, {\bar{\mathbf x}_i}^{1:M}
& \sim \mathcal N(\mathbf K_{i; NM}\mathbf K_{i; M}^{-1}\bar{\mathbf f}_i, \ \mathbf V_i) \nonumber\\
\bar{\mathbf f_i}\ |\ {\bar{\mathbf x}_i}^{1:M} & \sim \mathcal N(0, \mathbf K_{i; M})
\label{eq:spgp}
\end{align}

\noindent where $\mathbf K_{i; NM}$ is a $N \times M$ matrix with each
$(j, k)$ element given by the kernel function $k_i(\mathbf
x_{P_i}^{(j)}, {\mathbf {\bar x}_{i}}^{(k)})$. Similarly, $\mathbf
K_{i; M}$ is a $M \times M$ matrix where element $(j, k)$ is
$k_i({\mathbf {\bar x}_{i}}^{(j)}, {\mathbf {\bar x}_{i}}^{(k)})$.  It
is important to notice that each pseudo-input ${\bar{\mathbf
    X}_i}^{(d)}$, $d = 1, \dots, M$, has the same dimensionality as
$\mathbf X_{P_i}$. The motivation for this is that $\bar{\mathbf X}_i$
works as an alternative training set, with the original prior
predictive means and variances being recovered if $M = N$ and $\bar{\mathbf X}_i =
\mathbf X_{P_i}$.

Let $\mathbf k_{i; dM}$ be the $d^{th}$ row of $\mathbf K_{i; NM}$.
Matrix $\mathbf V_i$ is a diagonal matrix with entry
$v_{i; dd}$ given by $v_{i; dd} = k_i(\mathbf x_{P_i}^{(d)}, \mathbf x_{P_i}^{(d)}) - 
\mathbf k_{i; dM}^{\mathsf T}\mathbf K_{i; M}^{-1}\mathbf k_{i; dM}$.
This implies that all latent function values $\{f_i^{(1)}, \dots, f_i^{(N)}\}$
are conditionally independent.

\subsection{Pseudo-inputs: A Fully Bayesian Formulation}

The density function implied by (\ref{eq:spgp}) replaces the
standard Gaussian process prior. In the context of \cite{snelson:06},
input and output variables are observed, and as such Snelson and
Ghahramani optimize ${\mathbf {\bar x}_{i}}^{1:M}$ by maximizing the
marginal likelihood of the model.  This is practical but sometimes
prone to overfitting, since pseudo-inputs are in fact free parameters,
and the pseudo-inputs model is best seen as a variation of the
Gaussian process prior rather than an approximation to it
\citep{titsias:09}. 

In our setup, there is limited motivation to optimize the
pseudo-inputs since the inputs themselves are random variables.  For
instance, we show in the next section that the cost of sampling
pseudo-inputs is no greater than the cost of sampling latent
variables, while avoiding cumbersome optimization techniques to choose
pseudo-input values. Instead we put a prior on the pseudo-inputs and
extend the sampling procedure. By conditioning on the data, a good
placement for the pseudo-inputs can be learned, since $\mathbf
X_{P_i}$ and ${{\mathbf {\bar X}_i}}^{(d)}$ are dependent in the
posterior. This is illustrated by Figure \ref{fig:bspgp}. Moreover,
it naturally provides a protection against overfitting.

%Practical aspects need to be considered when choosing a family of
%pseudo-input priors. If one links to pseudo-inputs to actual inputs
%$\mathbf X_{P_i}^{1:N}$ in the prior, then computational and
%statistical issues arise.  For instance, making each pseudo-input
%${\bar{\mathbf X}_{i}}^{(m)}$ to be a deterministic function of all
%inputs is problematic since the whole matrix $\mathbf K_{i; M}$ will
%change every time we sample an element of any $\mathbf X_{P_i}^{(d)}$,
%$1 \le d \le N$. This would imply a cost of $\mathcal O(M^3N)$ for a
%single iteration of sampling, defeating the purpose of the
%pseudo-input formulation for all but very small $M$.

A simple choice of priors for pseudo-inputs is as follows:
each pseudo-input ${\bar{\mathbf X}_i}^{(d)}$, $d = 1, \dots, M$, is
given a $\mathcal N(\mu_i^{d}, \Sigma_i^{d})$ prior, independent of
all other random variables. A partially informative (empirical)
prior can be easily defined in the case where, for each $X_k$,
we have the freedom of choosing a particular indicator
$Y_q$ with fixed structural equation $Y_q = X_k + \epsilon_q$ (see
Section \ref{sec:identifiability}), implying $E[X_k] = E[Y_q]$. This
means if $X_k$ is a parent $X_i$, we set the respective entry in
$\mu_i^{d}$ (recall $\mu_i^{d}$ is a vector with an entry for every
parent of $X_i$) to the empirical mean of $Y_q$.  Each prior
covariance matrix $\Sigma_i^{d}$ is set to be diagonal with a common
variance.

Alternatively, we would like to spread the pseudo-inputs a priori:
other things being equal, pseudo-inputs that are too close to each can
be wasteful given their limited number. One prior, inspired by
space-filling designs from the experimental design literature
\citep{sant:will:notz:2003}, is
\begin{equation*}
p(\bar{\mathbf x}_i^{1:M}) \propto \det(\mathbf D_i)
\end{equation*}
\noindent the determinant of a kernel matrix $\mathbf D_i$. We use a
squared exponential covariance function with characteristic length
scale of $0.1$ \citep{rassmilliams:06}, and a ``nugget'' constant that
adds $10^{-4}$ to each diagonal term. This prior has support over a
$[-L, L]^{|\mathbf X_{P_i}|}$ hypercube. We set $L$ to be three times the
largest standard deviation of observed variables in the training data.
This is the pseudo-input prior we adopt in our experiments, where
we center all observed variables at their empirical means.

\subsection{Other Priors}

We adopt standard priors for the parametric components of this model:
independent Gaussians for each coefficient $\lambda_{ij}$, inverse
gamma priors for the variances of the error terms and a Dirichlet
prior for the distribution of the mixture indicators of the exogenous
variables.

\section{INFERENCE}
\label{sec:mcmc}

We use a Metropolis-Hastings scheme to sample from our space of latent
variables and parameters. Similarly to Gibbs sampling, we sample blocks
of random variables while conditioning on the remaining variables. When the
corresponding conditional distributions are canonical, we sample 
directly from them. Otherwise, we use mostly standard random walk proposals.

Conditioned on the latent variables, sampling the parameters of the
measurement model is identical to the case of classical Bayesian
linear regression. The same can be said of the sampling scheme for the
posterior variances of each $\zeta_i$. Sampling the mixture
distribution parameters for the exogenous variables is also identical
to the standard Bayesian case of Gaussian mixture models. Details are
described in the Appendix.

We describe the central stages of the sampler for the sparse model.
The sampler for the model with full Gaussian process priors is simpler and
analogous, and also described in the Appendix.

\subsection{Sampling Latent Functions}

In principle, one can analytically marginalize the pseudo-functions
${\bar{\mathbf f}_i}^{1:M}$.  However, keeping an explicit sample of
the pseudo-functions is advantageous when sampling latent variables
$X_i^{(d)}$, $d = 1, \dots, N$: for each child $X_c$ of $X_i$, only
the corresponding factor for the conditional density of $\mathbf
f_c^{(d)}$ needs to be computed (at a $\mathcal O(M)$ cost), since
function values are independent given latent parents and
pseudo-functions. This issue does not arise in the fully-observed case of
\cite{snelson:06}, who do marginalize the pseudo-functions.

Pseudo-functions and functions $\{{\bar{\mathbf f}_i}^{1:M}, \mathbf
f_i^{1:N}\}$ are jointly Gaussian given all other random variables and
data. The conditional distribution of
$\bar{\mathbf f}_i^{1:M}$ given everything, except itself {\it and}
$\{f_i^{(1)}, \dots, f_i^{(N)}\}$, is Gaussian with covariance matrix
\begin{equation*}
\bar{\mathbf S}_i \equiv (\mathbf K_{i; M}^{-1} + 
\mathbf K_{i; M}^{-1}\mathbf K_{i; NM}^\mathsf{T}
(\mathbf V_i^{-1} + \mathbf I/\upsilon_{\zeta_i})
\mathbf K_{i; NM}\mathbf K_{i; M}^{-1})^{-1}
\end{equation*}
where $\mathbf V_i$ is defined in Section \ref{sec:pseudo-review} and
$\mathbf I$ is a $M \times M$ identity matrix. The total cost of
computing this matrix is $\mathcal O(NM^2 + M^3) = \mathcal O(NM^2)$. 
The corresponding mean is
\begin{equation*}
\bar{\mathbf S}_i \times \mathbf K_{i; M}^{-1}\mathbf K_{i; NM}^\mathsf{T}
(\mathbf V_i^{-1} + \mathbf I/\upsilon_{\zeta_i})\mathbf x_i^{1:N}
\end{equation*}
\noindent where $\mathbf x_i^{1:N}$ is a column vector of length $N$.

Given that $\bar{\mathbf f}_i^{1:M}$ is sampled according to this
multivariate Gaussian, we can now sample $\{f_i^{(1)}, \dots,
f_i^{(N)}\}$ in parallel, since this becomes a mutually independent
set with univariate Gaussian marginals. The conditional variance of
$f_i^{(d)}$ is $v_i' \equiv 1 / (1 / v_{i; dd} + 1 /
\upsilon_{\zeta_i})$, where $v_{i; dd}$ is defined in Section
\ref{sec:pseudo-review}. The corresponding mean is $v_i'(f_\mu^{(d)} /
v_{i; dd} + x_i^{(d)} / \upsilon_{\zeta_i})$, where $f_\mu^{(d)} =
\mathbf k_{i;dM}\mathbf K_{i; M}^{-1}\bar{\mathbf f}_i$.

In Section \ref{sec:experiments}, we also sample from the posterior
distribution of the hyperparameters $\Theta_i$ of the kernel function
used by $\mathbf K_{i;M}$ and $\mathbf K_{i;NM}$. Plain Metropolis-Hastings
is used to sample these hyperparameters, using an uniform proposal in
$[\alpha \Theta_i, (1/\alpha)\Theta_i]$ for $0 < \alpha < 1$.

\subsection{Sampling Pseudo-inputs and Latent Variables}

We sample each pseudo-input $\bar {\mathbf x}_i^{(d)}$ one at a time,
$d = 1, 2, \dots, M$. Recall that $\bar {\mathbf x}_i^{(d)}$ is a
vector, with as many entries as the number of parents of $X_i$. In our
implementation, we propose all entries of the new $\bar {\mathbf
  x}_i^{(d)'}$ simultaneously using a Gaussian random walk proposal
centered at $\bar {\mathbf x}_i^{(d)'}$ with the same variance in each
dimension and no correlation structure.  For problems where the number
of parents of $X_i$ is larger than in our examples (i.e., four or more
parents), other proposals might be justified.

Let $\bar{\pi}_i^{(\backslash d)}(\bar {\mathbf x}_i^{(d)})$ be the conditional prior for 
$\bar {\mathbf x}_i^{(d)}$ given $\bar {\mathbf x}_i^{(\backslash d)}$,
where $(\backslash d) \equiv \{1, 2, \dots, d - 1, d + 1, \dots, M\}$.
Given a proposed $\bar {\mathbf x}_i^{(d)'}$, we accept the new value with
probability $\min \left\{1, {l_i(\bar{\mathbf x}_i^{(d)'})}/{l_i(\bar{\mathbf x}_i^{(d)})}\right\}$
where
\[
\begin{array}{rcl}
\displaystyle
l_i(\bar{\mathbf x}_i^{(d)}) &=&
 \bar{\pi}_i^{(\backslash d)}(\bar {\mathbf x}_i^{(d)'})
 \times p(\bar{f}_i^{(d)}\ |\ \bar{\mathbf f}_i^{(\backslash d)}, \bar{\mathbf x}_i)\\
&\times& \prod_{d = 1}^N v_{i;dd}^{-1/2}e^{-(f_i^{(d)} - \mathbf k_{i;dM}\mathbf K_{i;M}^{-1}\bar{\mathbf f}_i)^2/(2v_{i;dd})}\\
\end{array}
\]
and 
$p(\bar{f}_i^{(d)}\ |\ \bar{\mathbf f}_i^{(\backslash d)},
\bar{\mathbf x}_i)$ is the conditional density that follows from
Equation (\ref{eq:spgp}). Row vector $\mathbf k_{i;dM}$ is the $d^{th}$
row of matrix $\mathbf K_{i; NM}$. Fast submatrix updates of $\mathbf
K_{i;M}^{-1}$ and $\mathbf K_{i;NM}\mathbf K_{i;M}^{-1}$ are required in order
to calculate $l_i(\cdot)$ at a $\mathcal O(NM)$ cost,
which can be done by standard Cholesky updates \citep{seeger:04}. The
total cost is therefore $\mathcal O(NM^2)$ for a full sweep over all
pseudo-inputs.

The conditional density $p(\bar{f}_i^{(d)}\ |\ \bar{\mathbf
  f}_i^{(\backslash d)}, \bar{\mathbf x}_i)$ is known to be sharply
peaked for moderate sizes of $M$ (at the order of hundreds) \citep{titsias:09b},
which may cause mixing problems for the Markov chain.
One way to mitigate this effect is to also propose a value 
$\bar{f}_i^{(d)'}$ jointly with $\bar{\mathbf x}_i^{(d)'}$, which is possible
at no additional cost. We propose the pseudo-function using the conditional 
$p(\bar{f}_i^{(d)}\ |\ \bar{\mathbf  f}_i^{(\backslash d)}, \bar{\mathbf x}_i)$.
The Metropolis-Hastings acceptance probability for this variation is then simplified to
$\min \left\{1, l_i^0(\bar{\mathbf x}_i^{(d)'}) / l_i(\bar{\mathbf x}_i^{(d)})\right\}$,
where
\[
\begin{array}{rcl}
\displaystyle
l_i^0(\bar{\mathbf x}_i^{(d)}) &=&
\bar{\pi}_i^{(\backslash d)}(\bar {\mathbf x}_i^{(d)'})\\
&\times& \prod_{d = 1}^N v_{i;dd}^{-1/2}e^{-(f_i^{(d)} - \mathbf k_{i;dM}\mathbf K_{i;M}^{-1}\bar{\mathbf f}_i)^2/(2v_{i;dd})}\\
\end{array}
\]

Finally, consider the proposal for latent variables $X_i^{(d)}$.  For
each latent variable $X_i$, the set of latent variable instantiations
$\{X_i^{(1)}, X_i^{(2)}, \dots, X_i^{(N)}\}$ is mutually independent
given the remaining variables. We propose each new latent variable
value $x_i^{(d)'}$ in parallel, and accept or reject it based on a
Gaussian random walk proposal centered at the current value
$x_i^{(d)}$. We accept the move with probability
$\min \left\{1, h_{X_i}(x_i^{(d)'})/h_{X_i}(x_i^{(d)})\right\}$
where, if $X_i$ is not an exogenous variable in the graph, 
\[
\begin{array}{rcl}
\displaystyle
h_{X_i}(x_i^{(d)}) &=& e^{-(x_i^{(d)} - f_i^{(d)})^2 / (2\upsilon_{\zeta_i})} \\
&& \times \prod_{X_c \in \mathbf X_{C_i}}p(f_c^{(d)}\ |\ \bar{\mathbf f}_c, \bar{\mathbf x}_c, x_i^{(d)})\\
&& \times \prod_{Y_c \in \mathbf Y_{C_i}}p(y_c^{(d)}\ |\ \mathbf x_{P_c}^{(d)})
\end{array}
\]
\noindent where $\mathbf X_{C_i}$ is the set of latent children of
$X_i$ in the graph, and $\mathbf Y_{C_i}$ is the corresponding
set of observed children.

The conditional $p(f_c^{(d)}\ |\ \bar{\mathbf f}_c, \bar{\mathbf x}_c,
x_i^{(d)})$, which follows from 
(\ref{eq:spgp}), is a non-linear function of $x_i^{(d)}$, but crucially 
does not depend on any $x_i^{(\cdot)}$ variable except point $d$. The
evaluation of this factor costs $\mathcal O(M^2)$. As such, sampling
all latent values for $X_i$ takes $\mathcal O(NM^2)$.

The case where $X_i$ is an exogenous variable is analogous, given
that we also sample the mixture component indicators of such
variables.

\section{EXPERIMENTS}
\label{sec:experiments}

In this evaluation Section\footnote{MATLAB code to run all of our
  experiments is available at {\small \tt http://www.homepages.ucl.ac.uk/$\sim$ucgtrbd/}
{\small \tt code/gpsem.zip}}, 
we briefly illustrate the algorithm in a
synthetic study, followed by an empirical evaluation on how
identifiability matters in order to obtain an interpretable
distribution of latent variables. We end this section with a study
comparing the performance our model in predictive tasks against common
alternatives\footnote{Some implementation details: we used the squared
  exponential kernel function $k(\mathbf x_p, \mathbf x_q) = 
  a \exp(-\frac{1}{2b}|\mathbf x_p - \mathbf x_q|^2) + 10^{-4}\delta_{pq}$,
  where $\delta_{pq} = 1$ is $p = q$ and 0 otherwise.
  The hyperprior for $a$ is a mixture of a gamma $(1, 20)$ and a gamma
  $(10, 10)$ with equal probability each. The same (independent) prior is
  given to $b$. Variance parameters were given inverse gamma (2, 1) priors, and the
  linear coefficients were given Gaussian priors with a common large
  variance of 5. Exogenous latent variables were modeled as a mixture
  of five Gaussians where the mixture distribution is given a Dirichlet
  prior with parameter 10. Finally, for each latent $X_i$ variable we
  choose one of its indicators $Y_j$ and fix the corresponding edge
  coefficient to 1 and intercept to 0 to make the model identifiable.
  We perform $20,000$ MCMC iterations with a burn-in period of $2000$
  (only $6000$ iterations with $1000$ of burn-in for the non-sparse GPSEM-LV
  due to its high computational cost).
  Small variations in the priors for coefficients (using a variance of $10$)
  and variance parameters (using an inverse gamma $(2, 2)$), and a mixture of
  3 Gaussians instead of 5, were attempted with no significant differences
  between models.}.

\subsection{An Illustrative Synthetic Study}

\begin{figure*}[ht]
\begin{center}
\begin{tabular}{cccc}
\hspace{-0.10in}
\epsfig{file=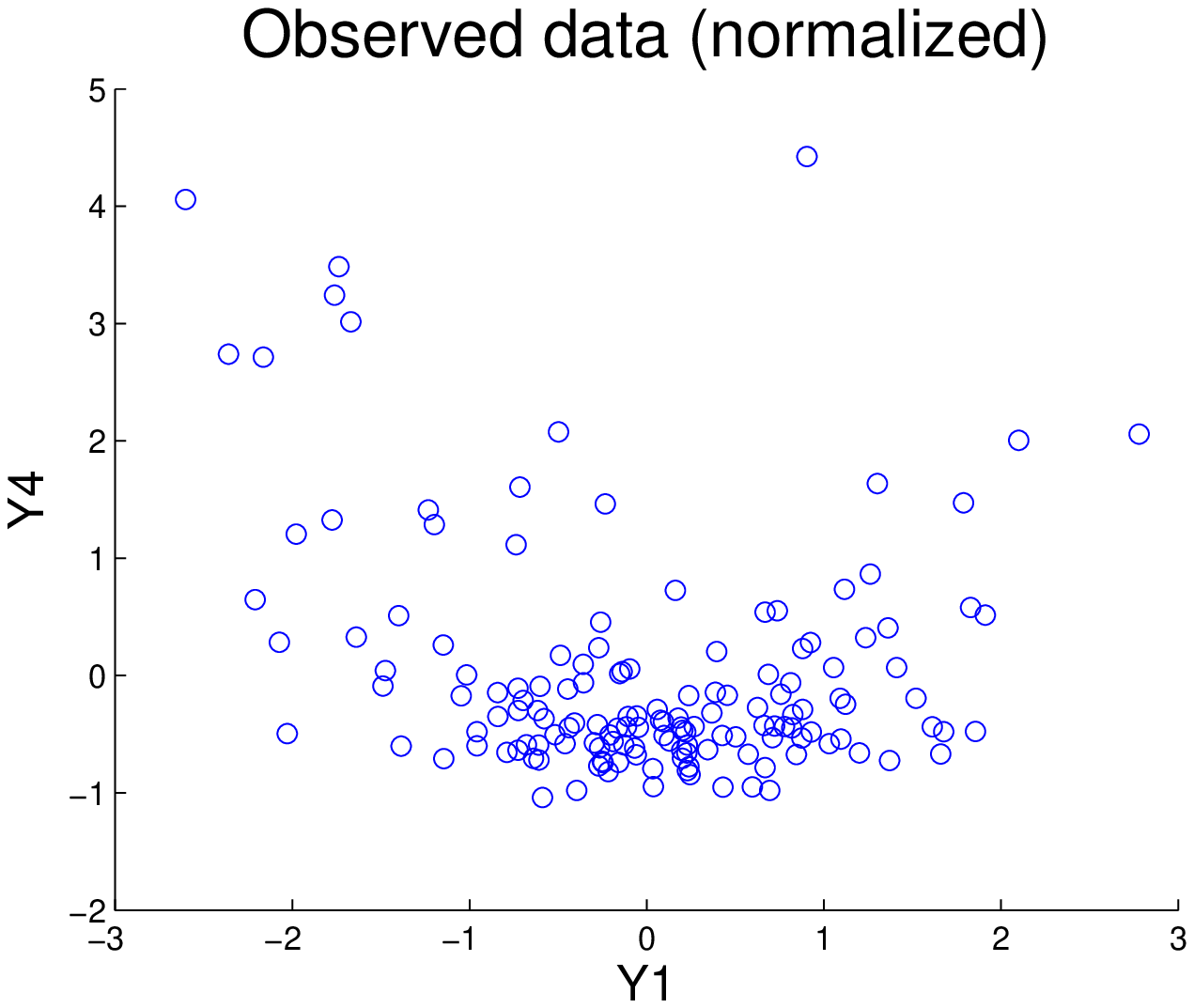,height=3.5cm} & \hspace{-0.4in}
\epsfig{file=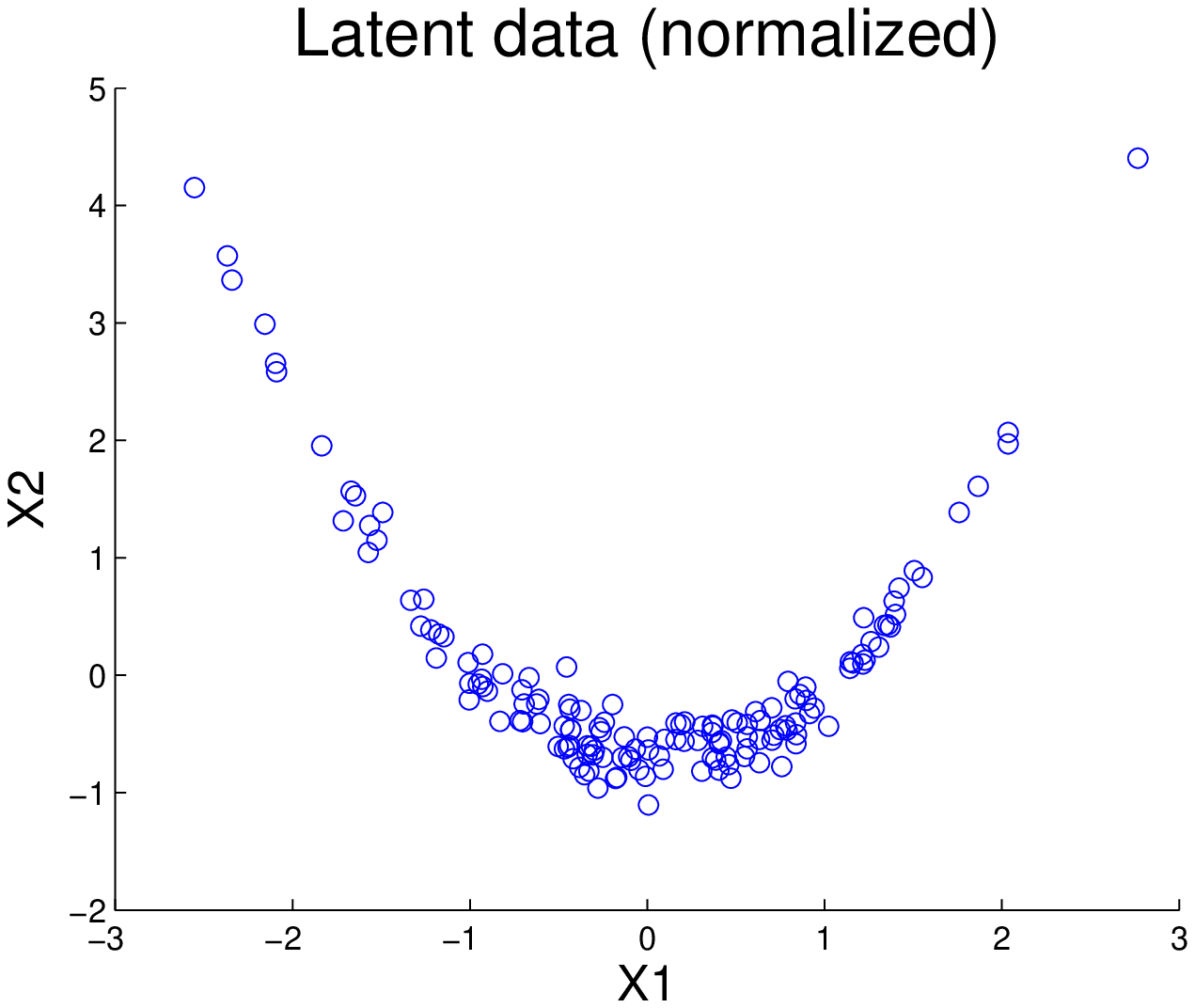,height=3.5cm} & \hspace{-0.4in}
\epsfig{file=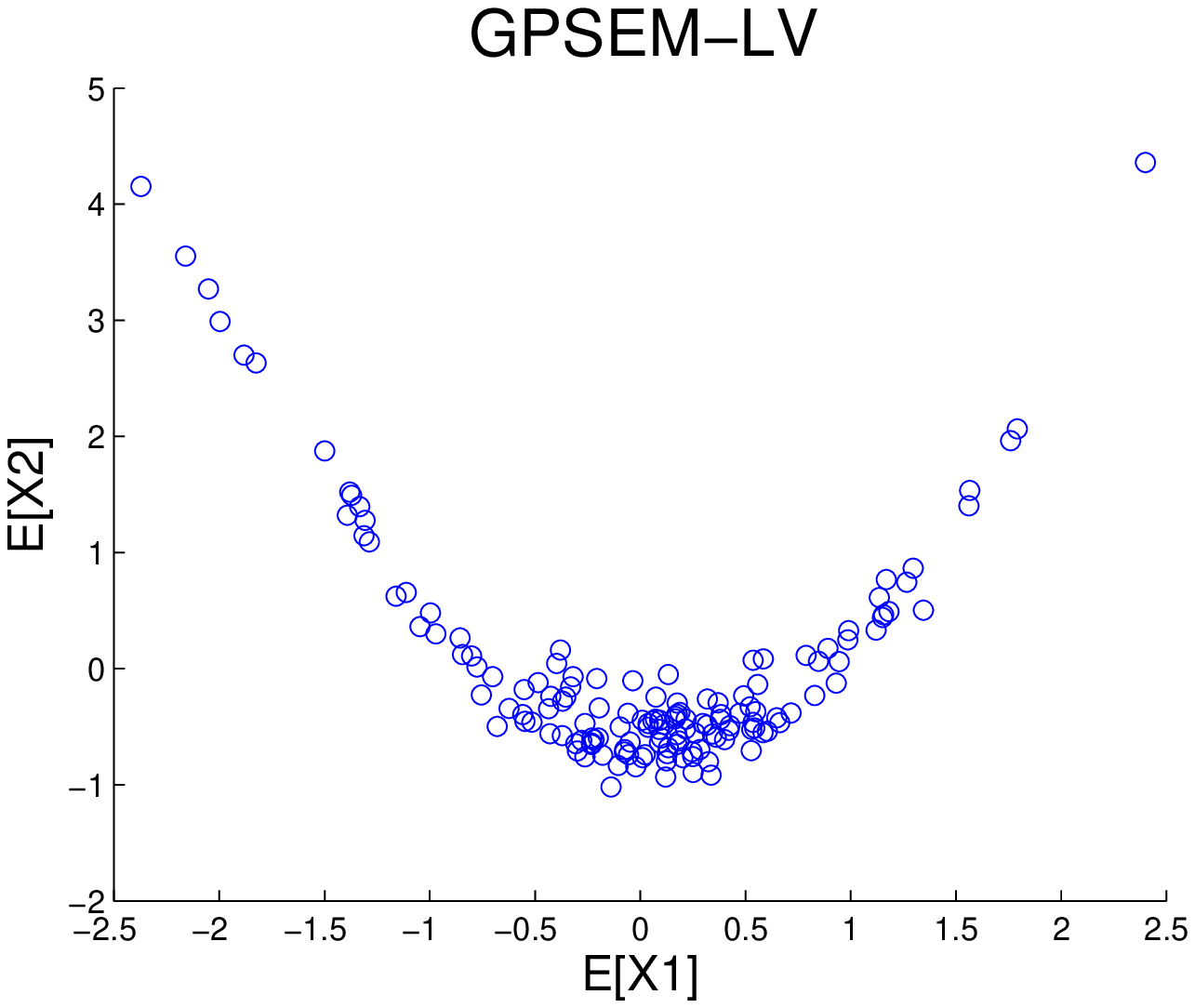,height=3.5cm} & \hspace{-0.35in}
\epsfig{file=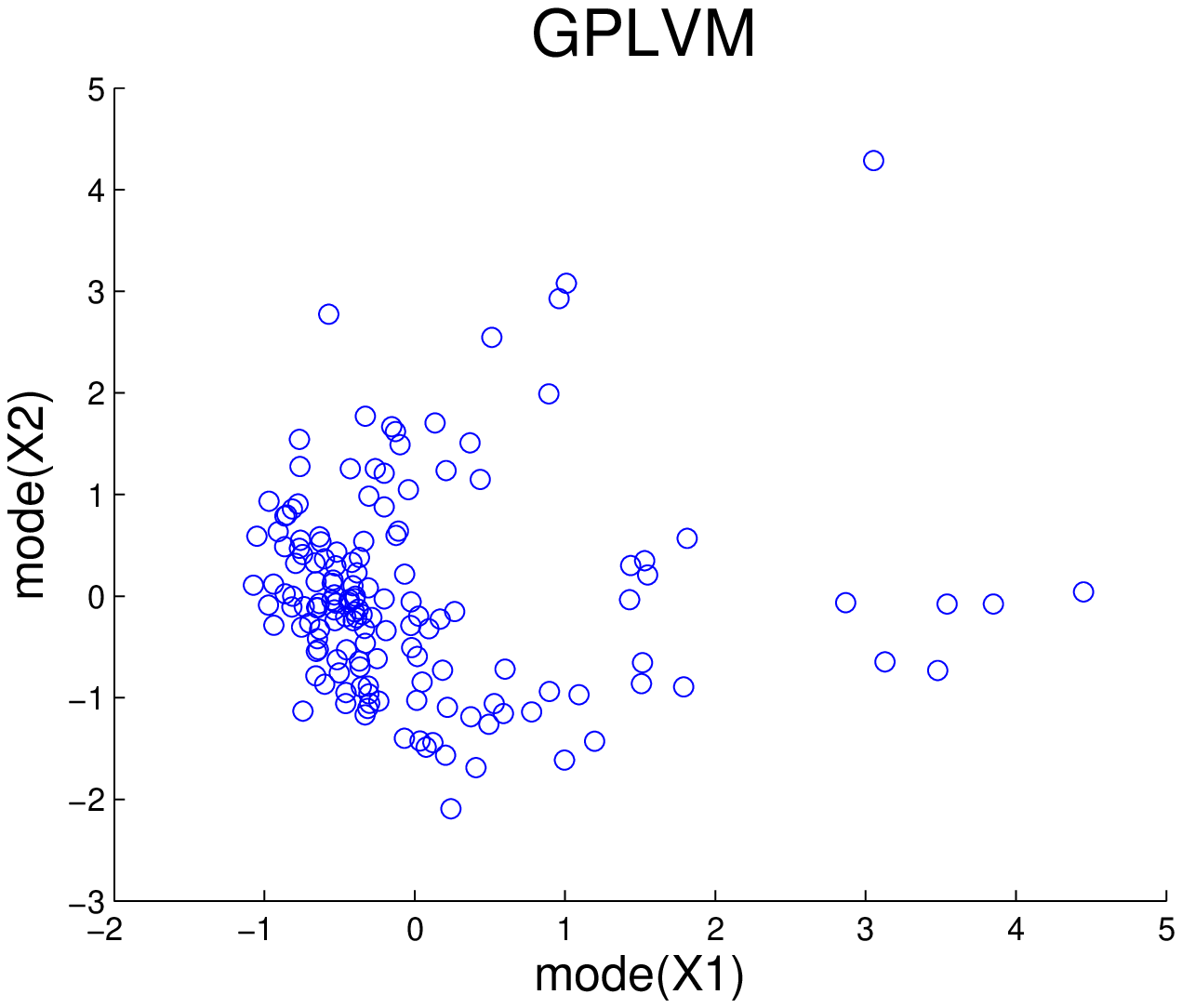,height=3.5cm} \\
(a) & (b) & (c) & (d)\\
\end{tabular}
\end{center}
\caption{(a) Plot of observed variables $Y_1$ and $Y_4$ generated by adding
standard Gaussian noise to two latent variables $X_1$ and $X_2$, where
$X_2 = 4X_1^2 + \zeta_2$, $\zeta_2$ also a standard Gaussian. 150 data
points were generated. (b) Plot of the corresponding latent variables,
which are not recorded in the data. (c) The posterior expected values
of the 150 latent variable pairs according to GPSEM-LV. (d) The
posterior modes of the 150 pairs according to GPLVM.}
\label{fig:synth}
\end{figure*}

We generated data from a model of two latent variables $(X_1, X_2)$ where $X_2 = 4X_1^2 + \zeta_2$,
$Y_i = X_1 + \epsilon_i$ for $i = 1, 2, 3$ and $Y_i = X_2 + 
\epsilon_i$, for $i = 4, 5, 6$. $X_1$ and all error terms follow standard
Gaussians. Given a sample of 150 points from this model, we set the
structural equations for $Y_1$ and $Y_4$ to have a zero intercept and
unit slope for identifiability purposes. Observed data for $Y_1$
against $Y_4$ is shown in Figure \ref{fig:synth}(a), which suggests a
noisy quadratic relationship (plotted in \ref{fig:synth}(b), but unknown to
the model). We run a GPSEM-LV model with 50
pseudo-inputs. The expected posterior value of each latent pair
$\{X_1^{(d)}, X_2^{(d)}\}$ for $d = 1, \dots, 150$ is plotted in
Figure \ref{fig:synth}(c). It is clear that we were able to reproduce
the original non-linear functional relationship given noisy data using
a pseudo-inputs model.

For comparison, the output of the Gaussian process latent variable model
\citep[GPLVM,][]{lawrence:05} with two hidden variables is shown in
Figure \ref{fig:synth}(d). GPLVM here assumes that the marginal
distribution of each latent variable is a standard Gaussian, but the
measurement model is nonparametric. In theory, GPLVM is as flexible as
GPSEM-LV in terms of representing observed joints. However, it does
not learn functional relationships among latent variables, which is
often of central interest in SEM applications
\citep{bol:89}. Moreover, since no marginal dependence among latent
variables is allowed, the model adapts itself to find (unidentifiable)
functional relationships between the exogenous latent variables of the
true model and the observables, analogous to the case illustrated by
Figure \ref{fig:simple-bollen}(b). As a result, despite GPLVM being able to
depict, as expected, some quadratic relationship (up to a rotation),
it is noisier than the one given by GPSEM-LV.

%The model can also be harder to interpret, since there is no explicit
%theory accounting for the identifiability of the model. In constrast,
%a consequence of SEM analysis is to force the data analyst to choose
%appropriate indicators that measure latent variables in a scale of
%interest. It is clear from Figure \ref{fig:synth}(d) that the inferred latent
%variables do not correspond to the independent latent variables $X_1$
%and $\zeta_2$ of the original model. The corresponding posterior expected
%values of such variables is depicted below:
%
%\begin{center}
%\epsfig{file=synth5.eps,height=3.5cm}
%\end{center}
%
%Although this might be a unfair comparison since GPSEM-LV uses knowledge
%about the sparsity structure of the measurement model, in SEM applications
%it is important to allow for the use of this information.

\subsection{MCMC and Identifiability}

We now explore the effect of enforcing identifiability constraints on the
MCMC procedure. We consider the dataset {\bf Consumer}, a
study\footnote{There was one latent variable marginally independent of
everything else. We eliminated it and its two indicators, as well as
the REC latent variable that had only 1 indicator.} with 333
university students in Greece \citep{bart:08}. The aim of the study was
to identify the factors that affect willingness to pay more to consume
environmentally friendly products. We selected 16 indicators of
environmental beliefs and attitudes, measuring a total of 4 hidden
variables. For simplicity, we will call these variables $X_1, \dots,
X_4$. The structure among latents is $X_1 \rightarrow X_2$, $X_1
\rightarrow X_3$, $X_2 \rightarrow X_3$, $X_2 \rightarrow X_4$.
Full details are given by \cite{bart:08}.

All observed variables have a single latent parent in the
corresponding DAG. As discussed in Section \ref{sec:identifiability},
the corresponding measurement model is identifiable by fixing the
structural equation for one indicator of each variable to have a zero
intercept and unit slope \citep{bart:08}. If the assumptions described
in the references of Section \ref{sec:identifiability} hold, then the
latent functions are also identifiable. We normalized the dataset
before running the MCMC inference algorithm.

An evaluation of the MCMC procedure is done by running and comparing 5
independent chains, each starting from a different point. Following
\cite{lee:07}, we evaluate convergence using the EPSR statistic
\citep{gelman:92}, which compares the variability of a given marginal
posterior within each chain and between chains. We calculate this
statistic for all latent variables $\{X_1, X_2, X_3, X_4\}$ across all
333 data points.

\begin{figure*}[ht]
\begin{center}
\begin{tabular}{cccc}
\hspace{-0.30in}
\epsfig{file=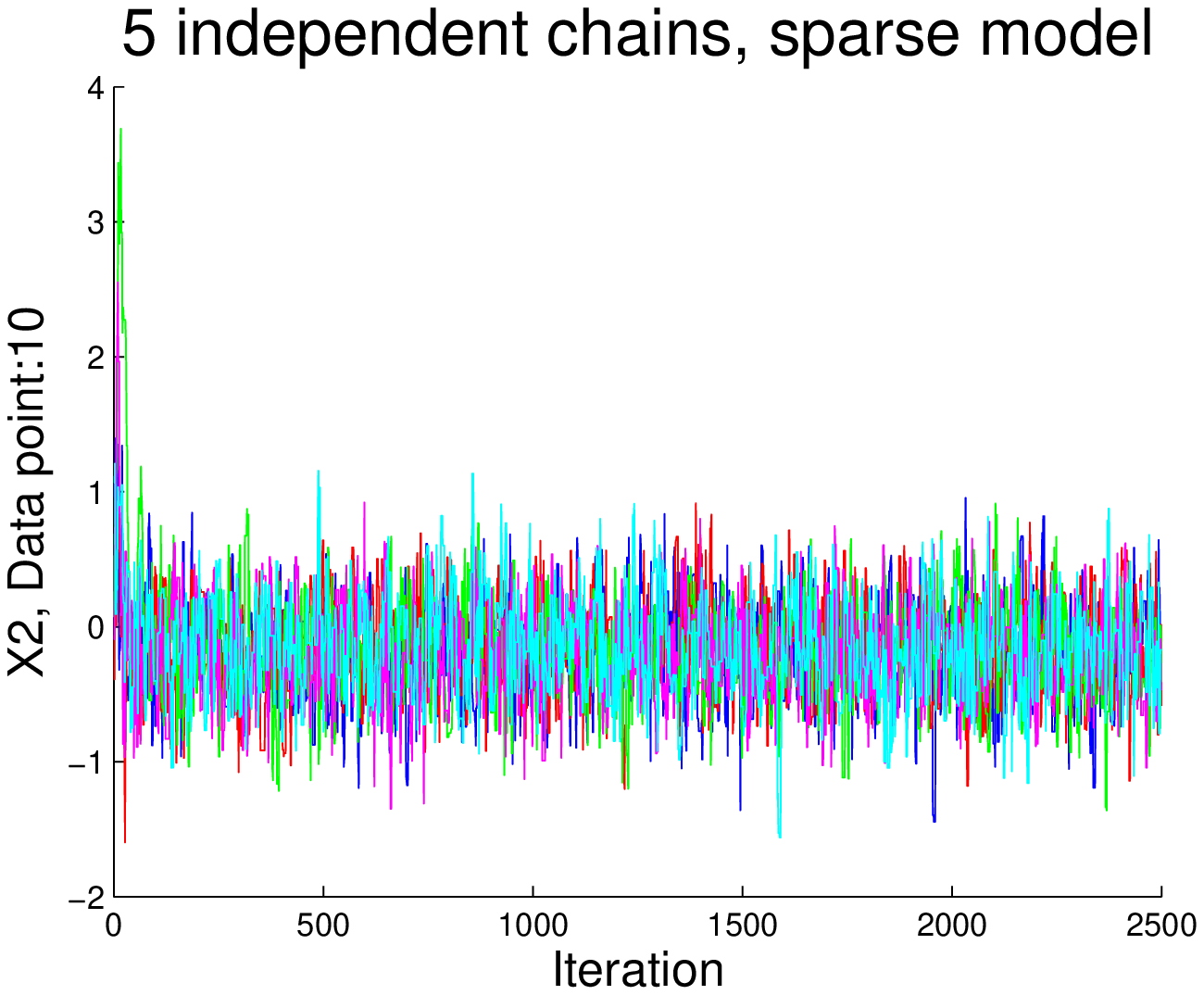,height=3.5cm,width=4.5cm}    & \hspace{-0.25in}
\epsfig{file=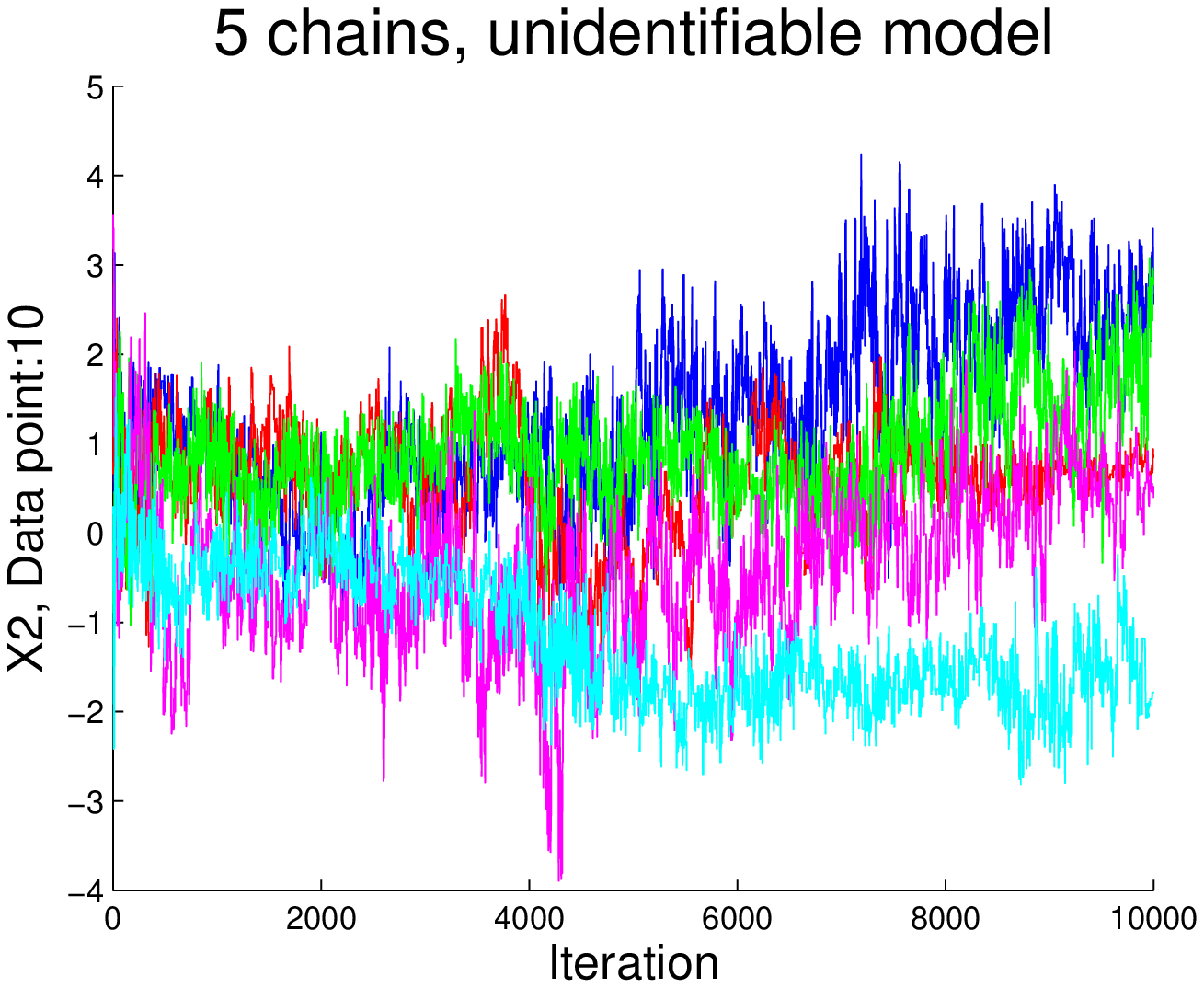,height=3.5cm,width=4.5cm} & \hspace{-0.25in}
\epsfig{file=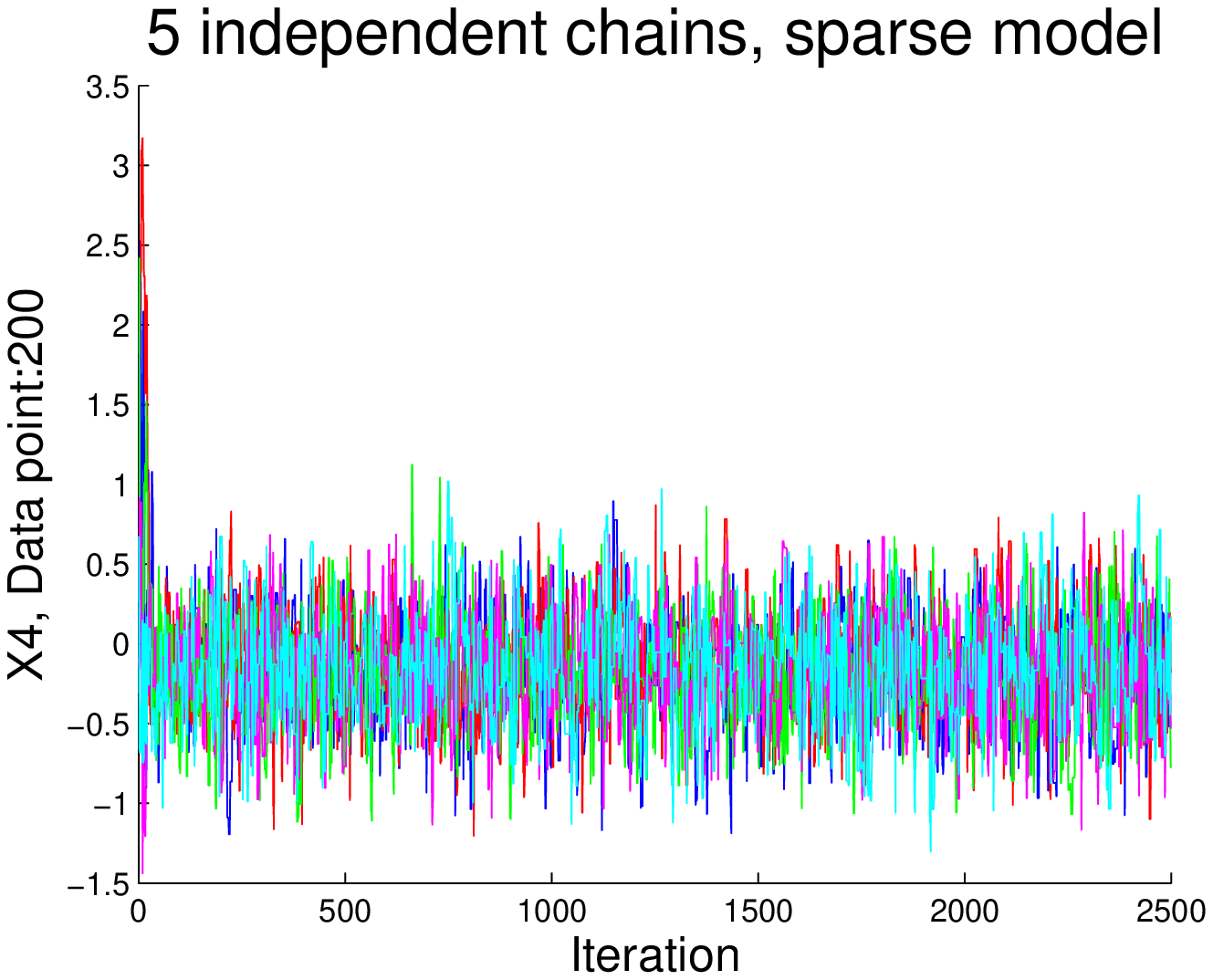,height=3.5cm,width=4.5cm}    & \hspace{-0.25in}
\epsfig{file=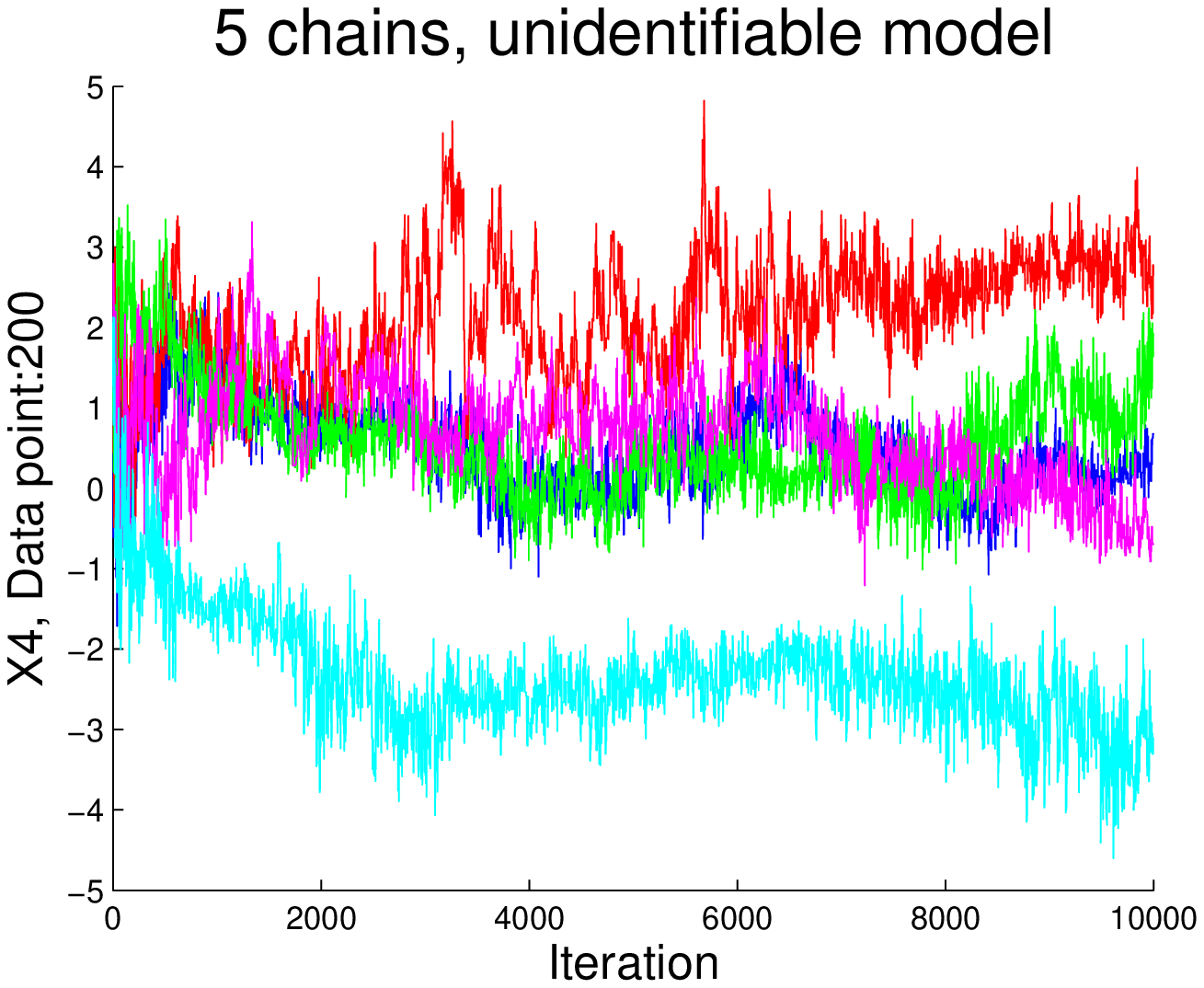,height=3.5cm,width=4.5cm}\\
\end{tabular}
\end{center}
\caption{An illustration of the behavior of independent chains for $X_2^{(10)}$ and $X_4^{(200)}$
using two models for the {\bf Consumer} data: the original (sparse)
model \citep{bart:08}; an (unidentifiable) alternative where the each
observed variable is an indicator of all latent variables. In the
unidentifiable model, there is no clear pattern across the independent
chains. Our model is robust to initialization, while the alternative
unidentifiable approach cannot be easily interpreted.}
\label{fig:5chains}
\end{figure*}

\begin{figure}[h]
\begin{center}
\hspace{-0.10in}
\epsfig{file=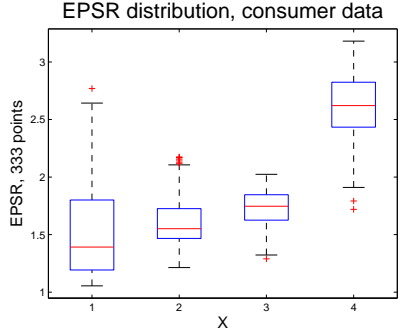,height=4.5cm}
\end{center}
\caption{Boxplots for the EPSR distribution across each of the 333 datapoints
of each latent variable. Boxes represent the distribution for
the non-sparse model. A value less than 1.1 is considered acceptable
evidence of convergence \citep{lee:07}, but this essentially never
happens. For the sparse model, all EPSR statistics were under 1.03.}
\label{fig:5chainsbox}
\end{figure}

A comparison is done against a variant of the model where the
measurement model is {\it not} sparse: instead, each observed variable
has all latent variables as parents, and no coefficients are fixed.
The differences are noticeable and illustrated in Figure
\ref{fig:5chains}. Box-plots of EPSR for the 4 latent variables are
shown in Figure \ref{fig:5chainsbox}. It is difficult to interpret or
trust an embedding that is strongly dependent on the initialization
procedure, as it is the case for the unidentifiable model. As
discussed by \cite{palomo:07}, identifiability might not be a
fundamental issue for Bayesian inference, but it is an important
practical issue in SEMs.

\subsection{Predictive Verification of the Sparse Model}

We evaluate how well the sparse GPSEM-LV model
performs compared against two parametric SEMs and GPLVM. The {\it
  linear} structural equation model is the SEM, where each latent
variable is given by a linear combination of its parents with additive
Gaussian noise. Latent variables without parents are given the same
mixture of Gaussians model as our GPSEM-LV implementation. The {\it
  quadratic} model includes all quadratic and linear terms, plus
first-order interactions, among the parents of any given latent
variable. This is perhaps the most common non-linear SEM used in
practice \citep{bollen:98,lee:07}. GPLVM is fit with 50 active points
and the {\tt rbf} kernel with automatic relevance determination
\citep{lawrence:05}. Each sparse GPSEM model uses $50$ pseudo-points.

We performed a 5-fold cross-validation study where the average 
predictive log-likelihood on the respective test sets is reported.
Three datasets are used. The first is the {\bf Consumer} dataset, 
described in the previous section. 

The second is the {\bf Abalone} data \citep{uci:07}, where we
postulate two latent variables, ``Size'' and ``Weight.'' 
{\it Size} has as indicators the length, diameter and height of each
abalone specimen, while {\it Weight} has as indicators the four weight
variables. We direct the relationship among latent variables as $Size
\rightarrow Weight$. 

The third is the {\bf Housing} dataset \citep{uci:07,harrison:78},
which includes indicators about features of suburbs in Boston that are
relevant for the housing market. Following the original study 
\cite[][Table IV]{harrison:78}, we postulate three latent
variables: ``Structural,'' corresponding to the structure of each
residence; ``Neighborhood,'' corresponding to an index of neighborhood
attractiveness; and ``Accessibility,'' corresponding to an index of
accessibility within Boston\footnote{The analysis by \cite[][Table
IV]{harrison:78} also included a fourth latent concept of ``Air
pollution,'' which we removed due to the absence of one of its
indicators in the electronic data file that is available.}.  The
corresponding 11 non-binary observed variables that are associated
with the given latent concepts are used as indicators. The
``Neighborhood'' concept was refined into two, ``Neighborhood I'' and
``Neighborhood II'' due to the fact that three of its original
indicators have very similar (and highly skewed) marginal
distributions, which were very dissimilar from the others\footnote{The
final set of indicators, using the nomenclature of the UCI repository
documentation file, is as follows: ``Structural'' has as indicators
$RM$ and $AGE$; ``Neighborhood I'' has as indicators $CRIM$, $ZN$ and
$B$; ``Neighborhood II'' has as indicators $INDUS$, $TAX$, $PTRATIO$
and $LSTAT$; ``Accessibility'' has as indicators $DIS$ and $RAD$. See
\citep{uci:07} for detailed information about these indicators. Following
Harrison and Rubinfield, we log-transformed some of the variables:
$INDUS$, $DIS$, $RAD$ and $TAX$.}. The
structure among latent variables is given by a fully connected network
directed according to the order $\{$Accessibility, Structural,
Neighborhood II, Neighborhood I$\}$. \cite{harrison:78} provide full
details on the meaning of the indicators.
We note that it is well known that the {\bf Housing} dataset poses stability
problems to density estimation due to discontinuities in the variable
$RAD$, one of the indicators of accessibility \citep{friedman:00b}. In
order to get more stable results, we use a subset of the data (374 points)
where $RAD < 24$.

\begin{figure}[h]
\begin{center}
\hspace{-0.10in}
\begin{tabular}{cc}
\hspace{-0.2in}
\epsfig{file=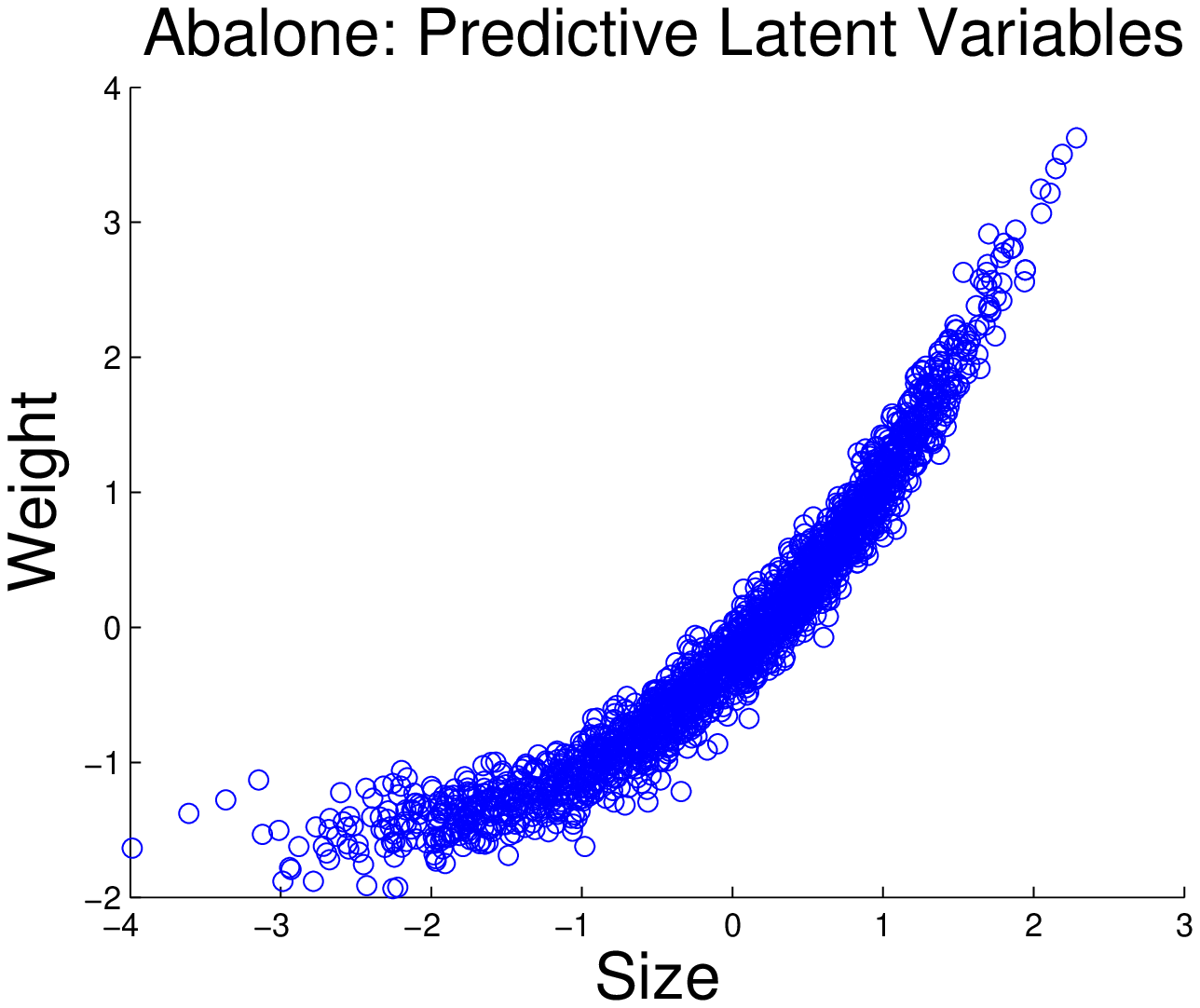,height=3.2cm} &
\hspace{-0.2in}
\epsfig{file=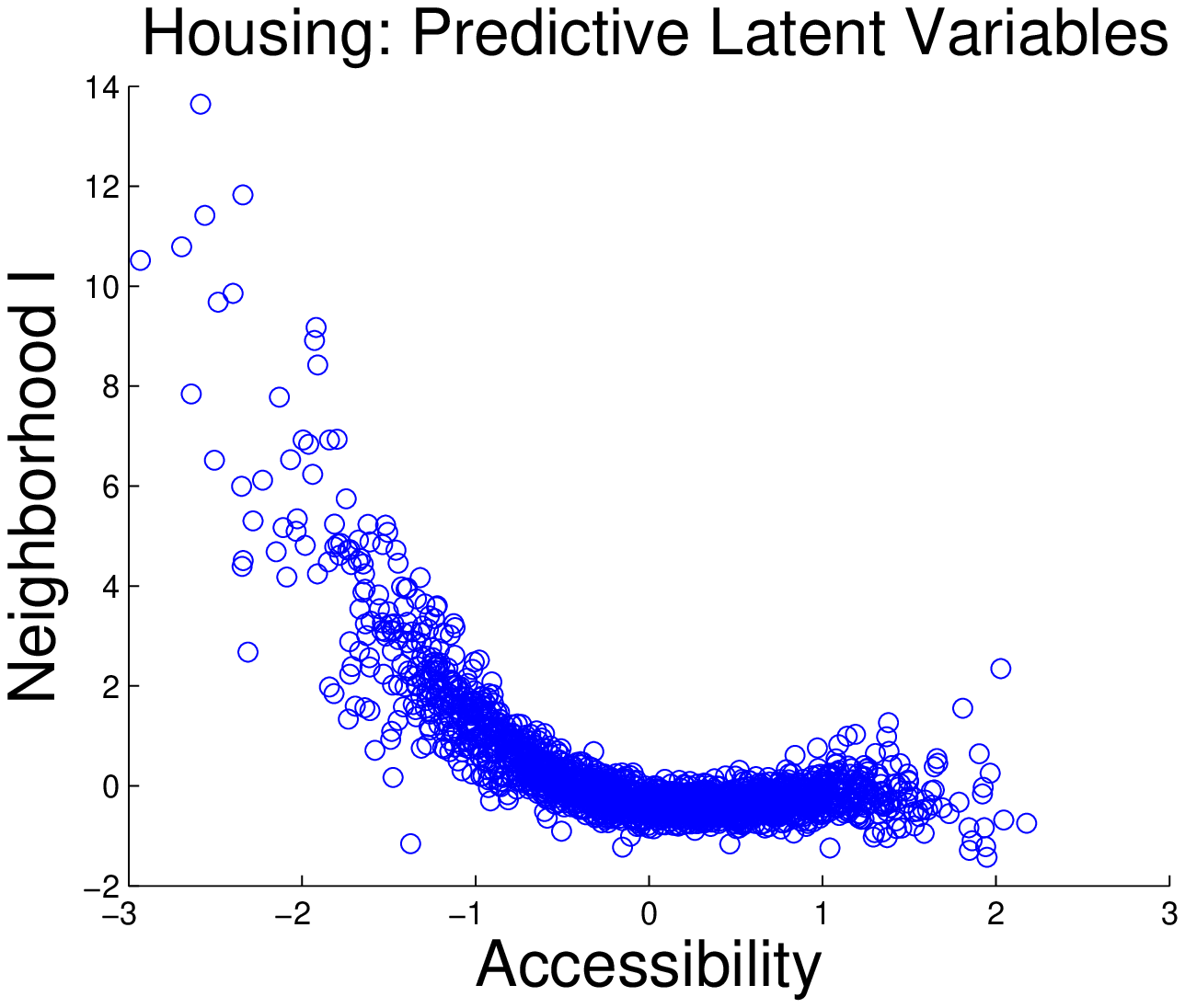,height=3.2cm}
\end{tabular}
\end{center}
\caption{Scatterplots of 2000 fantasy samples taken from the predictive distributions
of sparse GPSEM-LV models. In contrast, GPLVM would generate spherical Gaussians.}
\label{fig:pred}
\end{figure}

The need for non-linear SEMs is well-illustrated by Figure \ref{fig:pred},
where fantasy samples of latent variables are generated from the predictive
distributions of two models.

We also evaluate how the non-sparse GPSEM-LV behaves compared to the
sparse alternative. Notice that while {\bf Consumer} and {\bf Housing}
have each approximately 300 training points in each cross-validation fold,
{\bf Abalone} has over 3000 points. For the non-sparse GPSEM, we subsampled
all of {\bf Abalone} training folds down to 300 samples.

Results are presented in Table \ref{tab:llik}. Each dataset was chosen
to represent a particular type of problem. The data in {\bf Consumer}
is highly linear. In particular, it is important to point out that the
GPSEM-LV model is able to behave as a standard structural equation
model if necessary, while the quadratic polynomial model shows some
overfitting. The {\bf Abalone} study is known for having clear
functional relationships among variables, as also discussed by
\cite{friedman:00b}. In this case, there is a substantial difference
between the non-linear models and the linear one, although GPLVM seems
suboptimal in this scenario where observed variables can be easily
clustered into groups. Finally, functional relationships
among variables in {\bf Housing} are not as clear
\citep{friedman:00b}, with multimodal residuals. GPSEM still shows an
advantage, but all SEMs are suboptimal compared to GPLVM. One
explanation is that the DAG on which the models rely is not adequate.
Structure learning might be necessary to make the most out of
nonparametric SEMs.

\begin{table*}[t]
\small{
\begin{center}
\begin{tabular}{|r|ll|lll|ll|lll|ll|lll|}
\hline
 & \multicolumn{5}{c|}{\bf Consumer} & \multicolumn{5}{c|}{\bf Abalone} & \multicolumn{5}{c|}{\bf Housing}\\
\hline
         & GPS  \hspace{-0.15in} & GP \hspace{-0.00in} & LIN    \hspace{-0.15in} & QDR    \hspace{-0.10in} & GPL
         & GPS  \hspace{-0.15in} & GP \hspace{-0.00in} & LIN    \hspace{-0.15in} & QDR    \hspace{-0.10in} & GPL
         & GPS  \hspace{-0.15in} & GP \hspace{-0.00in} & LIN    \hspace{-0.15in} & QDR    \hspace{-0.10in} & GPL\\
\hline
Fold 1 & -20.66  \hspace{-0.15in} & -21.17  \hspace{-0.00in} & -20.67  \hspace{-0.15in}  & -21.20  \hspace{-0.10in} & -22.11  & 
          -1.96  \hspace{-0.15in} &  -2.08  \hspace{-0.00in} &  -2.75  \hspace{-0.15in}  &  -2.00  \hspace{-0.10in} &  -3.04  & 
         -13.92  \hspace{-0.15in} & -14.10  \hspace{-0.00in} & -14.46  \hspace{-0.15in}  & -14.11  \hspace{-0.10in} & -11.94  \\
Fold 2 & -21.03  \hspace{-0.15in} & -21.15  \hspace{-0.00in} & -21.06  \hspace{-0.15in}  & -21.08  \hspace{-0.10in} & -22.22  & 
          -1.90  \hspace{-0.15in} &  -2.97  \hspace{-0.00in} &  -2.52  \hspace{-0.15in}  &  -1.92  \hspace{-0.10in} &  -3.41  & 
         -15.07  \hspace{-0.15in} & -17.70  \hspace{-0.00in} & -16.20  \hspace{-0.15in}  & -15.12  \hspace{-0.10in} & -12.98  \\
Fold 3 & -20.86  \hspace{-0.15in} & -20.88  \hspace{-0.00in} & -20.84  \hspace{-0.15in}  & -20.90  \hspace{-0.10in} & -22.33  & 
          -1.91  \hspace{-0.15in} &  -5.50  \hspace{-0.00in} &  -2.54  \hspace{-0.15in}  &  -1.93  \hspace{-0.10in} &  -3.65  & 
         -13.66  \hspace{-0.15in} & -15.75  \hspace{-0.00in} & -14.86  \hspace{-0.15in}  & -14.69  \hspace{-0.10in} & -12.58  \\
Fold 4 & -20.79  \hspace{-0.15in} & -21.09  \hspace{-0.00in} & -20.78  \hspace{-0.15in}  & -20.93  \hspace{-0.10in} & -22.03  &
          -1.77  \hspace{-0.15in} &  -2.96  \hspace{-0.00in} &  -2.30  \hspace{-0.15in}  &  -1.80  \hspace{-0.10in} &  -3.40  & 
         -13.30  \hspace{-0.15in} & -15.98  \hspace{-0.00in} & -14.05  \hspace{-0.15in}  & -13.90  \hspace{-0.10in} & -12.84  \\
Fold 5 & -21.26  \hspace{-0.15in} & -21.76  \hspace{-0.00in} & -21.27  \hspace{-0.15in}  & -21.75  \hspace{-0.10in} & -22.72  & 
          -3.85  \hspace{-0.15in} &  -4.56  \hspace{-0.00in} &  -4.67  \hspace{-0.15in}  &  -3.84  \hspace{-0.10in} &  -4.80  & 
         -13.80  \hspace{-0.15in} & -14.46  \hspace{-0.00in} & -14.67  \hspace{-0.15in}  & -13.71  \hspace{-0.10in} & -11.87  \\
\hline
\end{tabular}
\end{center}
}
\caption{Average predictive log-likelihood in a 5-fold cross-validation setup.
  The five methods are the GPSEM-LV model with 50 pseudo-inputs (GPS), GPSEM-LV with
  standard Gaussian process priors (GP), the linear and quadratic
  structural equation models (LIN and QDR) and the Gaussian process latent variable
  model (GPL) of \cite{lawrence:05}, a nonparametric factor analysis model.
  For {\bf Abalone}, GP uses a subsample of the training data. The p-values given by a paired
  Wilcoxon signed-rank test, measuring the significance of positive differences between sparse GPSEM-LV and the quadratic
  model, are 0.03 (for {\bf Consumer}), 0.34 ({\bf Abalone}) and 0.09 ({\bf Housing}).}
\label{tab:llik}
\end{table*}

Although results suggest that the sparse model behaved better that the
non-sparse one (which was true of some cases found by Snelson and
Ghahramani, 2006, due to heteroscedasticity effects), such results
should be interpreted with care. {\bf Abalone} had to be subsampled in
the non-sparse case. Mixing is harder in the non-sparse
model since all datapoints $\{X_i^{(1)}, \dots, X_i^{(N)}\}$ are
dependent. While we believe that with larger sample sizes and denser
latent structures the non-sparse model should be the best, large
sample sizes are too expensive to process and, in many SEM
applications, latent variables have very few parents.

It is also important to emphasize that the wallclock sampling time
for the non-sparse model was an order of magnitude larger than the sparse
case with $M = 50$ $-$ even considering that 3000 training points 
were used by the sparse model in the {\bf Abalone} experiment, against
300 points by the non-sparse alternative.
%\footnote{The code was implemented
%in MATLAB. A fast implementation in C is likely to magnify the differences:
%Cholesky computations are the main bottleneck, already
%treated efficiently by MATLAB, but looping over 3000 training points
%against the 300 of the undersampled model would be more efficient in C.}.

\section{RELATED WORK}
\label{sec:related}

Non-linear factor analysis has been studied for decades in the
psychometrics literature\footnote{Another instance of the ``whatever
  you do, somebody in psychometrics already did it long before'' law:
  http://www.stat.columbia.edu/$\sim$cook/movabletype/archives/
  2009/01/a\_longstanding.html}. A review is provided by
\cite{yalcin:01}.  However, most of the classic work is based on
simple parametric models.  A modern approach based on Gaussian
processes is the Gaussian process latent variable model of
\cite{lawrence:05}. By construction, factor analysis cannot be used in
applications where one is interested in learning functions relating
latent variables, such as in causal inference. For embedding, factor
analysis is easier to use and more robust to model misspecification
than SEM analysis. Conversely, it does not benefit from well-specified
structures and might be harder to interpret. \cite{bol:89} discusses
the interplay between factor analysis and SEM. Practical non-linear
structural equation models are discussed by \cite{lee:07}, but none of
such approaches rely on nonparametric methods. Gaussian processes
latent structures appear mostly in the context of dynamical systems
(e.g., \cite{ko:09}). However, the connection is typically among data
points only, not among variables within a data point, where on-line
filtering is the target application.

\section{CONCLUSION}
\label{sec:conclusion}

The goal of graphical modeling is to exploit the structure of
real-world problems, but the latent structure is often ignored.  We
introduced a new nonparametric approach for SEMs by extending a
sparse Gaussian process prior as a fully Bayesian procedure.
Although a standard MCMC algorithm worked reasonably well, it is
possible as future work to study ways of improving mixing times. This
can be particularly relevant in extensions to ordinal variables, where
the sampling of thresholds will likely make mixing more difficult.
Since the bottleneck of the procedure is the sampling of the
pseudo-inputs, one might consider a hybrid approach where a subset of
the pseudo-inputs is fixed and determined prior to sampling using a cheap heuristic.
New ways of deciding pseudo-input locations based on a given
measurement model will be required. Evaluation with larger datasets
(at least a few hundred variables) remains an open problem.  Finally,
finding ways of determining the graphical structure is also a
promising area of research.

\section*{Acknowledgements}

We thank Patrick Hoyer for several relevant discussions concerning
the results of Section \ref{sec:identifiability}, and Irini Moustaki
for the consumer data.

\small{
\bibliography{rbas}
}

\section*{APPENDIX A: FURTHER MCMC DETAILS}

We use a MCMC sampler to draw all variables of interest from the posterior
distribution of a GPSEM model. Let $M$ denote the number of pseudo-inputs 
per latent function $f_i(\cdot)$, $N$ be the sample size, $V$ the 
number of latent variables and $K$ the common number of Gaussian mixture
components for each exogenous latent variable.

The sampler is a standard Metropolis-Hastings procedure with block
sampling: random variables are divided into blocks, where we sample
each block conditioning on the current values of the remaining blocks.

We consider both the non-sparse and sparse variations of GPSEM.
The blocks are as follows for the non-sparse GPSEM:
\begin{itemize}
\item the linear coefficients for the structural equation of each
observed variable $Y_j$: $\{\lambda_{j0}\} \cup \Lambda_j$
\item the conditional variance for the structural equation of each
observed variable $Y_j$: $\upsilon_{\epsilon_j}$
\item the $d$-th instantiation of each latent variable $X_i$,
$x_i^{(d)}$;
\item the set of latent function values $\{f_i^{(1)}, \dots, f_i^{(N)}\}$
for each particular endogenous latent variable $X_i$
\item the conditional variance for the structural equation of each
latent variable $X_i$: $\upsilon_{\zeta_i}$
\item the set of latent mixture component indicators $\{z_i^{(1)}, \dots
z_i^{(N)}\}$ for each particular exogenous latent variable $X_i$
\item the set of means $\{\mu_{i1}, \dots, \mu_{iK}\}$ for the mixture
components of each particular exogenous latent variable $X_i$
\item the set of variances $\{v_{i1}, \dots, v_{iK}\}$ for the mixture
components of each particular exogenous latent variable $X_i$
\item mixture distribution $\pi_i$ corresponding to the
probability over mixture components for exogenous latent variable
$X_i$
\end{itemize}

The blocks for the sparse model are similar, except that
\begin{itemize}
\item all instantiations of a given latent variable
$x_i^{(d)}$, for $d = 1, 2, \dots, N$,
are mutually independent conditioned on the functions, pseudo-inputs
and pseudo-functions. As such, they can be treated as a single block
of size $N$, where all elements are sampled in parallel;
\item the $d$-th instantiation of each pseudo-input
$\bar{\mathbf x}_i^{(d)}$ for $d = 1, 2, \dots, M$
\item all instantiations of latent functions and pseudo-latents functions
$\{f_i^{(1)}, \dots, f_i^{(N)}, \bar{f}_i^{(1)}, \dots, \bar{f}_i^{(M)}\}$
for any particular $X_i$ are conditionally multivariate Gaussian and can be
sampled together
\end{itemize}

We adopt the convention that, for any particular step described in the
following procedure, any random variable that is not explicitly
mentioned should be considered fixed at the current sampled value.
Moreover, any density function that depends on such implicit variables
uses the respective implicit values.

Our implementation uses code for submatrix Cholesky updates from the library
provided by \cite{seeger:04}.

\subsection*{The measurement model}

The measurement model can be integrated out in principle, if we adopt
a conjugate normal-inverse gamma prior for the linear regression of
observed variables $\mathbf Y$ on $\mathbf X$. However, we opted for a
non-conjugate prior in order to evaluate the convergence of the
sampler when this marginalization cannot be done (as in alternative
models with non-Gaussian error terms).

Given the latent variables, the corresponding conditional
distributions for the measurement model parameters boil down to
standard Bayesian linear regression posteriors. In our
Metropolis-Hastings scheme, our proposals correspond to such
conditionals, as in Gibbs sampling (and therefore have an acceptance
probability of 1).

Let $\mathbf X_{P_j}$ be the parents of observed variable $Y_j$ in the
graph and let the $d$-th instantiation of the corresponding regression
input be $\tilde{\mathbf x}_{P_j}^{(d)} \equiv [\mathbf
x_{P_j}^\mathsf{T} 1]^\mathsf{T}$. Let each cofficient $\lambda_{jk}$
have an independent Gaussian prior with mean zero and variance $u$.
Conditioned on the error variance $\upsilon_{\epsilon_j}$, the
posterior distribution of the vector $[\lambda_{j1}, \dots,
\lambda_{j|\mathbf X_{P_j}|}, \lambda_{j0}]^\mathsf{T}$ is
multivariate Gaussian with covariance $\mathbf S_j \equiv (\sum_{d =
  1}^N \tilde{\mathbf x}_{P_j}^{(d)} \tilde{\mathbf
  x}_{P_j}^{(d)\mathsf{T}} + \mathbf I / u)^{-1}$ and mean $\mathbf
S_j\sum_{d = 1}^N \tilde{\mathbf x}_{P_j}^{(d)}y_j^{(d)}$, where
$\mathbf I$ is a $(|\mathbf X_{P_j}| + 1) \times (|\mathbf X_{P_j}| +
1)$ identity matrix.

The derivation for the case where some coefficients $\lambda_{jk}$ are
fixed to constants is analogous.

For a fixed set of linear coefficients $\{\lambda_{j0}\} \cup
\Lambda_j$, we now sample the conditional variance
$\upsilon_{\epsilon_j}$. Let this variance have a inverse gamma prior
$(a, b)$. Its conditional distribution is an inverse gamma $(a', b')$,
where $a' = a + N / 2$, $b' = b + \sum_{d = 1}^N (\hat{e}_j^{(d)})^2
/2$, and $\hat{e}_j^{(d)} \equiv y_j^{(d)} - \lambda_{j0} -
\Lambda_j^\mathsf{T}\mathbf{x}_{P_j}^{(d)}$.

\subsection*{The structural model: non-sparse GPSEM}

For all $i = 1, 2, \dots, V$ and $d = 1, 2, \dots, N$, we propose each
new latent variable value $x_i^{(d)'}$ individually, and accept or
reject it based on a Gaussian random walk proposal centered at the
current value $x_i^{(d)}$. We accept the move with probability
\begin{equation*}
\displaystyle
\min \left\{1, \frac{g_{X_i}(x_i^{(d)'})}{g_{X_i}(x_i^{(d)})}\right\}
\end{equation*}
where, if $X_i$ is not an exogenous variable in the graph, 
\begin{equation}
\begin{array}{rcl}
\displaystyle
g_{X_i}(x_i^{(d)}) &=& e^{-(x_i^{(d)} - f_i^{(d)})^2 / (2\upsilon_{\zeta_i})}\\
& \times & \prod_{X_c \in \mathbf X_{C_i}}p(f_c^{(d)}\ |\ f_c^{(\backslash d)})\\
& \times & \prod_{Y_c \in \mathbf X_{Y_i}}p(y_c^{(d)}\ |\ \mathbf X_{P_c}^{(d)})
\end{array}
\label{eq:sample_x}
\end{equation}

Recall that $f_i(\cdot)$ is a function of the parents $\mathbf
X_{P_i}$ of $X_i$ in the graph. The $d$-th instantiation of such
parents assume the value $\mathbf x_{P_i}^{(d)}$. We use $f_i^{(d)}$
as a shorthand notation for $f_i(\mathbf x_{P_i}^{(d)})$. Morever, let
$\mathbf X_{C_i}$ denote the latent children of $X_i$ in the graph.
The symbol $f_c^{(\backslash d)}$ refers to the respective function
values taken by $f_c$ in data points $\{1, 2, \dots, d - 1, d + 1,
\dots, N\}$. Function $p(f_c^{(d)}\ |\ f_c^{(\backslash d)})$ is the
conditional density of $f_c^{(d)}$ given $f_c^{(\backslash d)}$,
according to the Gaussian process prior. The evaluation of this factor
costs $\mathcal O(N^2)$ using standard submatrix Cholesky updates
\citep{seeger:04}. As such, sampling all latent values for $X_i$
takes $\mathcal O(N^3)$.

Finally, $\mathbf X_{Y_i}$ denotes the observed children of $X_i$, and
function $p(y_c^{(d)}\ |\ \mathbf X_{P_c}^{(d)})$ is the corresponding
density of observed child $Y_c$ evaluated at $y_c^{(d)}$, given its
parents (which includes $X_i^{(d)}$) and (implicit) measurement model
parameters. This factor can be dropped if $y_c^{(d)}$ is missing.

If variable $X_i$ is an exogenous variable, then the factor
$e^{-(x_i^{(d)} - f_i^{(d)})^2/(2\upsilon_{\zeta_i})}$ gets substituted by 
\begin{equation*}
\displaystyle
e^{-\frac{1}{2}\left(x_i^{(d)} - \mu_{iz_i^{(d)}}\right)^2/v_{iz_i^{(d)}}}
\end{equation*}
\noindent where $z_i^{(d)}$ is the latent mixture indicator for the
marginal mixture of Gaussians model for $X_i$, with means $\{\mu_{i1}, \dots,
\mu_{iK}\}$ and variances $\{v_{i1}, \dots, v_{iK}\}$.

Given all latent variables, latent function values $\{f_i^{(1)}, \dots,
f_i^{(N)}\}$ are multivariate Gaussian with covariance matrix
\begin{equation*}
\mathbf S_{f_i} \equiv (\mathbf K_i^{-1} + \mathbf I / \upsilon_{\zeta_i})^{-1}
\end{equation*}
where $\mathbf K_i$ is the corresponding kernel matrix and $\mathbf I$
is a $N \times N$ identity matrix. The respective mean is given by
$\mathbf S_{f_i} \mathbf x_i^{(1:N)} / \upsilon_{\zeta_i}$, where
$\mathbf x_i^{(1:N)} \equiv [x_i^{(1)} \dots x_i^{(N)}]^\mathsf{T}$.
This operation costs $\mathcal O(N^3)$.  We sample from this
conditional as in a standard Gibbs update.

Sampling each latent conditional variance $\upsilon_{\zeta_i}$ can
also be done by sampling from its conditional. Let
$\upsilon_{\zeta_i}$ have an inverse gamma prior $(a_\zeta,
b_\zeta)$. The conditional distribution for this variance given all
other random variables is inverse gamma $(a'_\zeta, b'_\zeta)$, where
$a'_\zeta = a_\zeta + N / 2$ and
$b'_\zeta = b_\zeta + \sum_{d = 1}^N (x_i^{(d)} - f_i^{(d)})^2/2$.

We are left with sampling the mixture model parameters that correspond
to the marginal distributions of the exogenous latent variables. Once
we condition on the latent variables, this is completely standard. If
each mixture mean parameter $\mu_{ij}$ is given an independent
Gaussian prior with mean zero and variance $v_\pi$, its conditional
given the remaining variables is also Gaussian with variance $v_\pi'
\equiv 1/(1 / v_\pi + |Z_{ij}| / v_{ij})$, where $Z_{ij}$ is the
subset of $1, 2, \dots, N$ such that $d \in Z_{ij}$ if and only if $z_i^{(d)}
= j$. The corresponding mean is given by $v_\pi'\sum_{d \in Z_{ij}}
x_i^{(d)}/v_{ij}$.  If each mixture variance parameter $v_{ij}$ is
given an inverse gamma prior $(a_\pi, b_\pi)$, its conditional is an
inverse gamma $(a_\pi', b_\pi')$, where $a_\pi' = a_\pi + |Z_{ij}| /
2$, and $b_\pi' = b_\pi + \sum_{d \in Z_{ij}}(x_i^{(d)} -
\mu_{ij})^2/2$. The conditional probability 
$P(z_i^{(d)} = j\ |\ \text{everything else})$ is
proportional to $\sum_{t = 1}^N v_{ij}^{-1/2}e^{-(x_i^{(t)} -
  \mu_{ij})^2/(2v_{ij})}$.  Finally, given a Dirichlet prior
distribution $(\alpha_1, \dots, \alpha_K)$ for each $\pi_i$, its
conditional is also Dirichlet with parameter vector $(\alpha_1 +
|Z_{i1}|, \dots, \alpha_K + |Z_{iK}|)$.

\section*{APPENDIX B: A NOTE ON DIRECTIONALITY DETECTION}

The assumption of linearity of the measurement model is not only a
matter of convenience. In SEM applications, observed variables are
carefully chosen to represent different aspects of latent concepts of
interest and often have a single latent parent. As such, it is
plausible that children of a particular latent variable are different
noisy linear transformations of the target latent variable. This
differs from other applications of latent variable Gaussian process
models such as those introduced by \cite{lawrence:05}, where
measurements are not designed to explicitly account for target latent
variables of interest. Moreover, this linearity condition has important 
implications on distinguishing among candidate models.

\subsection*{Implications for Model Selection}

We assumed that the DAG $\mathcal G$ is given.  A detailed discussion
of model selection is left as future work.  Instead, we discuss some
theoretical aspects of a very particular but important structural
feature that will serve as a building block to more general model
selection procedures, in the spirit of \cite{hoyer:08}: determining
sufficient conditions for the subproblem of detecting edge
directionality from the data. Given a measurement model for two latent
variables $X_1$ and $X_2$, we need to establish conditions in which we
can test whether the only correct latent structure is $X_1 \rightarrow
X_2$, $X_2 \rightarrow X_1$, the disconnected structure, or either
directionality. The results of \cite{hoyer:08} can be extended to the
latent variable case by exploiting the conditions of identifiability
discussed in Section \ref{sec:identifiability} as follows.

Our sufficient conditions are a weaker set of assumptions than that of
\cite{sil:06}. We assume that $X_1$ has at least two observable children
which are not children of $X_2$ and vice-versa. Call these sets
$\{Y_1, Y_1'\}$ and $\{Y_2, Y_2'\}$, respectively.  Assume all error
terms ($\{\epsilon_i\} \cup \{\zeta_i\}$) are
non-Gaussian\footnote{Variations where $\epsilon_i$ and latent error
terms are allowed to be Gaussian, as in our original model description
are also possible and will be treated in the future.}. The variance of
all error terms is assumed to be nonzero. As in \cite{hoyer:08}, we also
assume $X_1$ and $X_2$ are unconfounded.

To test whether the model where $X_1$ and $X_2$ are independent
becomes easy in this case: the independence model entails that (say)
$Y_1$ and $Y_2$ are marginally independent. This can be tested using
the nonparametric marginal independence test of \cite{gretton:07}.

For the nontrivial case where latent variables are dependent, the
results of Section \ref{sec:identifiability} imply that the
measurement model of $\{X_1 \rightarrow Y_1, X_1 \rightarrow Y_1'\}$
is identifiable up to the scale and sign of the latent variables,
including the marginal distributions of $\epsilon_1$ and
$\epsilon_{1'}$. An analogous result applies to $\{X_2, Y_2, Y_2'\}$.

Since the measurement model $\{Y_1, Y_1', Y_2, Y_2'\}$ of $\{X_1,
X_2\}$ is identifiable, assume without loss of generality that the
linear coefficients corresponding to $X_1 \rightarrow Y_1$ and $X_2
\rightarrow Y_2$ are fixed to 1, i.e., $Y_1 = X_1 + \epsilon_i$ and
$Y_2 = X_2 + \epsilon_2$. Also from Section \ref{sec:identifiability},
it follows that the distribution of $\{X_1, X_2\}$ can be identified
under very general conditions. The main result of \cite{hoyer:08} can
then be directly applied. That is, data generated by a model $X_2 =
f(X_1) + \eta_2$, with $\eta_2$ being non-Gaussian and independent of
$X_1$, cannot be represented by an analogous generative model $X_1 =
g(X_2) + \eta_1$ except in some particular cases that are ruled out as
implausible.

\subsection*{Practical Testing}

The test for comparing $X_1 \rightarrow X_2$ against $X_2 \rightarrow
X_1$ in \cite{hoyer:08} can be modified to our context as follows:
we cannot regress $X_2$ on $X_1$ and estimate the residuals $\zeta_2$
since $X_1$ and $X_2$ are latent. However, we can do a error-in-variables
regression of $Y_2$ on $Y_1$ using $Y_1'$ and $Y_2'$ as instrumental variables 
\citep{carroll:04}: this means we find a function $h(\cdot)$ such that
$Y_2 = h(X) + r$ and $Y_1 = X + w$, for non-Gaussian latent variables
$r, w$ and $X$. We then calculate the estimated residuals $r$ of this
regression, and test whether such residuals are independent of $Y_1$
\citep{gretton:07}. If this is true, then we have no evidence to
discard the hypothesis $X_1 \rightarrow X_2$.

The justification for this process is that, if the true model is
indeed $X_2 = f(X_1) + \eta_2$, then $h(\cdot) = f(\cdot)$ and $r =
\epsilon_2 + \eta_2$ in the limit of infinite data, since the
error-in-variables regression model is identifiable in our case
\citep{carroll:04}, with $X = X_1$ being a consequence of deconvolving
$Y_1$ and $\epsilon_1$.  By this result, $r$ will be independent of
$Y_1$. However, if the opposite holds ($X_1 \leftarrow X_2$) then, as
in \citep{hoyer:08}, the residual is not in general independent of
$Y_1$: given $X_1$ ($=X$), there is a d-connecting path $Y_2
\leftarrow X_2 \rightarrow X_1 \leftarrow \eta_1$ \citep{pearl:00},
and $r$ will be a function of $\eta_1$, which is dependent on $Y_1$.
This is analogous to \citep{hoyer:08}, but using a different family of
regression techniques.

Error-in-variables regression is a special case of the Gaussian
process SEM. The main practical difficulty on using GPSEM with the
pseudo-inputs approximation in this case is that such pseudo-inputs
formulation implies a heteroscedastic regression model
\citep{snelson:06}. One has either to use the GPSEM formulation
without pseudo-inputs, or a model linear in the parameters but with an
explicit, finite, basis dictionary on the input space.

\end{document}